%% file: main.tex
\documentclass[10pt,twocolumn,letterpaper]{article}

\usepackage[pagenumbers]{iccv} 


\definecolor{iccvblue}{rgb}{0.21,0.49,0.74}

\usepackage{graphicx}
\usepackage{booktabs}

\usepackage{colortbl}
\usepackage{multirow}

\usepackage{wrapfig}
\usepackage{makecell} 
\usepackage{marvosym}

\usepackage[pagebackref,breaklinks,colorlinks,allcolors=iccvblue]{hyperref}

\usepackage[accsupp]{axessibility}  

\def\paperID{4674} 
\def\confName{ICCV}
\def\confYear{2025}

\title{MMOne: Representing Multiple Modalities in One Scene}

\author{
Zhifeng Gu \qquad Bing Wang$^{\dagger}$ \\
Spatial Intelligence Group, The Hong Kong Polytechnic University  \\
{\tt\small zhifeng.gu@connect.polyu.hk, bingwang@polyu.edu.hk}
}

\begin{document}

\twocolumn[{%
	\renewcommand\twocolumn[1][]{#1}%
	\maketitle
	\vspace{-2em}
	\includegraphics[width=1.\linewidth]{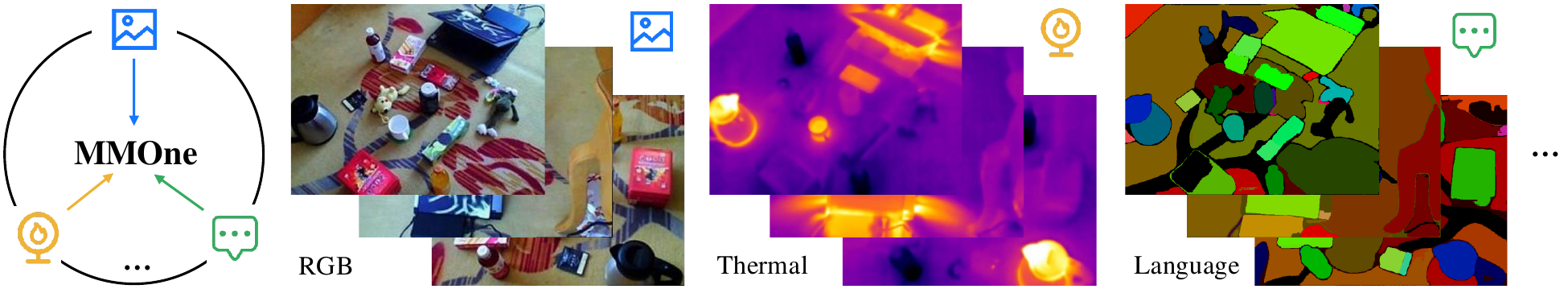}
	
    \captionof{figure}{\textbf{MMOne Overview.} MMOne is a general framework designed to represent multiple modalities in one scene. By disentangling multimodal information based on the inherent differences among modalities, we achieve enhanced performance across all modalities. }
	\label{fig:teaser}
	\vspace{12pt}
}]

\maketitle

\let\thefootnote\relax\footnotetext{$^{\dagger}$: Corresponding author.}

\input{sec/0_abstract}

\section{Introduction}
\label{sec:intro}

\input{sec/1_introduction}

\section{Related Work}
\label{sec:related_work}

\input{sec/2_related_work}

\section{Analysis of Multimodal Representation}
\label{sec:analysis}

\input{sec/3_analysis}

\section{Proposed Method}
\label{sec:method}

\input{sec/3_method}

\section{Experiments}
\label{sec:experiments}

\input{sec/4_experiments}

\section{Conclusion}

\input{sec/5_conclusion}

\vspace{6pt}
\noindent\textbf{Acknowledgement.} This work was jointly supported by the Young Scientists Fund of the Research Grants Council of Hong Kong (25206524) and the National Natural Science Foundation of China (42301520).

\clearpage

{
    \small
    \bibliographystyle{ieeenat_fullname}
    \bibliography{main}
}

\clearpage


\input{sec/6_appendix}

\end{document}

%% file: sec/0_abstract.tex
\begin{abstract}

Humans perceive the world through multimodal cues to understand and interact with the environment. Learning a scene representation for multiple modalities enhances comprehension of the physical world. However, modality conflicts, arising from inherent distinctions among different modalities, present two critical challenges: property disparity and granularity disparity. To address these challenges, we propose a general framework, \textbf{MMOne}, to represent multiple modalities in one scene, which can be readily extended to additional modalities. Specifically, a modality modeling module with a novel modality indicator is proposed to capture the unique properties of each modality. Additionally, we design a multimodal decomposition mechanism to separate multi-modal Gaussians into single-modal Gaussians based on modality differences. We address the essential distinctions among modalities by disentangling multimodal information into shared and modality-specific components, resulting in a more compact and efficient multimodal scene representation. Extensive experiments demonstrate that our method consistently enhances the representation capability for each modality and is scalable to additional modalities. The code is available at \href{https://github.com/Neal2020GitHub/MMOne}{https://github.com/Neal2020GitHub/MMOne}.


\vspace{-0.4cm}

\end{abstract}

%% file: sec/1_introduction.tex
The 3D physical world inherently comprises multiple modalities, such as vision, thermal, and language. Humans perceive these modalities to understand and interact with the environment \cite{wallace2004development}. Scene representation is a fundamental cognitive tool for humans to comprehend the world and is crucial for spatial cognition \cite{epstein2019scene, intraub2012rethinking}. Integrating information from different modalities into a multimodal scene representation enables better comprehension of the physical world, especially in complex environments \cite{duan2022multimodal, zhang2020multimodal}.

Scene representation has evolved from explicit representations \cite{guo2020deep} to implicit representations \cite{sitzmann2020implicit}, owing to their flexibility in geometry and appearance modeling through continuous parameterization. Neural Radiance Fields (NeRF) \cite{mildenhall2021nerf, tancik2023nerfstudio} and 3D Gaussian Splatting (3DGS) \cite{kerbl20233d} have demonstrated significant potential in various real-world applications, such as autonomous driving \cite{yan2024street, zhou2024drivinggaussian} and robotic manipulation \cite{shorinwa2024splatmover}, due to their capabilities for high-fidelity scene reconstruction and real-time rendering.

Beyond RGB, modalities such as language \cite{kerr2023lerf, liu2023weakly, qin2024langsplat, zhou2024feature, shi2024language} and thermal \cite{hassan2024thermonerf, lu2025thermalgaussian} have been incorporated into scene representation. Current works typically employ additional feature vectors to model modality-specific properties and render these modalities similarly to RGB. Follow-up studies incorporate modality-specific designs, such as learning distinctive features \cite{peng2025d, wu2024opengaussian}, to improve the modality representation ability. However, these approaches are limited to specific modalities and do not address the fundamental differences among different modalities. More critically, they use the same set of Gaussians to represent all modalities, which contradicts the varying levels of granularity of different modalities. These factors motivate us to consider a core problem: how to address the essential differences among modalities when representing multiple modalities simultaneously? This is inherently challenging as different modalities exhibit distinct properties and granularities.

\textbf{Modality conflicts}, which derived from the inherent differences among modalities, lead to two critical challenges when representing multiple modalities simultaneously. The \textbf{first} challenge is property disparity, which refers to the inherent distinctions in the characteristics of data from various modalities. The dimensionality and physical properties of diverse modalities vary (\textit{e.g.}, language representation requires higher dimensionality than RGB), which requires carefully designed modality representation to model. The \textbf{second} challenge is granularity disparity, which denotes the differences in the level of granularity at which information is represented across modalities (\textit{e.g.}, thermal exhibits coarser-grained than RGB). This issue is exacerbated when different modalities share the same geometry, resulting in redundant scene representations.

To tackle these challenges, we propose a general framework, \textbf{MMOne}, to represent multiple modalities in one scene, which can be readily extended to additional modalities. Specifically, a \textbf{modality modeling module} with modality-specific features and a novel modality indicator is proposed to capture the unique properties of each modality. The modality indicator also functions as a ``switch'' to selectively deactivate certain modalities during rendering. Additionally, we design a \textbf{multimodal decomposition mechanism} to separate multi-modal Gaussians into single-modal Gaussians based on modality differences, accommodating the varying levels of granularity of different modalities.

Our method addresses the essential distinctions among modalities by disentangling multimodal information into shared and modality-specific components, resulting in a more compact and efficient multimodal scene representation. Overall, our contributions are as follows:

\begin{itemize}
\setlength{\itemsep}{0pt}
\setlength{\parsep}{0pt}
\setlength{\parskip}{0pt}

    \item We propose MMOne, a general framework to represent multiple modalities in one scene, which can be extended to accommodate additional modalities.

    \item We design a modality modeling module to capture the modality-specific properties and a multimodal decomposition mechanism to separate multimodal information into shared and modality-specific components. 

    \item Extensive experimental evaluations on multiple benchmarks and datasets demonstrate the effectiveness and scalability of our method, consistently enhancing the representation capability for each modality. 
    
\end{itemize}

%% file: sec/2_related_work.tex
\subsection{Single-Modal Scene Representation}

Single-modal scene representation models the RGB modality using Neural Radiance Fields (NeRF) \cite{mildenhall2021nerf, gao2022nerf, barron2021mip, tancik2023nerfstudio} and 3D Gaussian Splatting (3DGS) \cite{kerbl20233d, chen2025survey3dgaussiansplatting}, which have significantly enhanced image quality in novel view synthesis. Notably, 3DGS has been applied to various domains, such as dynamic scenes \cite{wu20244d, xie2024physgaussian}, human avatars \cite{qian2024gaussianavatars, qian20243dgs}, and autonomous driving \cite{yan2024street, zhou2024drivinggaussian}, attributed to its explicit nature and high rendering efficiency. Numerous enhancements have been proposed to improve the rendering quality and efficiency of the RGB modality \cite{mallick2024taming, liu2025efficientgs, fang2024mini, fang2024mini2}. Beyond the RGB modality, thermal image has growing research potential and promising applications across various domains \cite{osornio2018recent, jalil2019visible, haque2020illuminating, dubail2022privacy, saputra2021graph}. Thermal3D-GS \cite{chen2024thermal3d} represents the thermal modality with 3DGS by introducing two neural networks to model thermal infrared physical characteristics.

\subsection{Dual-Modal Scene Representation}

Dual-modal scene representation simultaneously represents RGB with other modalities, including language, depth, thermal, and tactility \cite{dou2024tactile, swann2024touch}. For the language modality, LERF \cite{kerr2023lerf} and 3D-OVS \cite{liu2023weakly} build NeRF-based language fields with vision language models \cite{radford2021learning, caron2021emerging} to support open-vocabulary queries. GS-based methods extend 3DGS to jointly represent the RGB and language modalities, by introducing an extra language feature vector to each Gaussian \cite{qin2024langsplat, zhou2024feature, shi2024language, wu2024opengaussian, qu2024goi}. Feature dimensionality reduction is often applied to adapt the high-dimensional feature space of the language modality \cite{qin2024langsplat, shi2024language, wu2024opengaussian} , and a smoothed semantic indicator is introduced to disentangle the language rasterization \cite{peng2025d}. For the depth modality, depth or normal priors are incorporated to encourage the Gaussians to align with object surfaces \cite{huang20242d, turkulainen2024dnsplatter, yu2024gaussian, chen2024pgsr}, thus enhancing the accuracy of the scene geometry modeling. For the thermal modality, NeRF-based \cite{poggi2022cross, hassan2024thermonerf, ozer2024exploring} and GS-based \cite{lu2025thermalgaussian} methods have extended NeRF and 3DGS to represent RGB and thermal modalities simultaneously, by applying a thermal feature vector (\textit{e.g.}, spherical harmonics \cite{lu2025thermalgaussian}). However, these works typically use the same opacity and the same set of Gaussians to represent different modalities, which is suboptimal due to their distinct properties and granularities. Additionally, their modality-specific designs hinder their capability to accommodate more modalities.

\subsection{Multi-Modal Scene Representation}

Representing multiple ($>$2) modalities simultaneously is a field that remains underexplored. Recently, GLS \cite{qiu2024gls} and LangSurf \cite{li2024langsurf} utilize depth cues to align RGB-language Gaussians with object surfaces, facilitating precise segmentation with text queries. However, these methods are essentially ``dual-modal'', as they do not incorporate new properties into each Gaussian. Furthermore, they use the same Gaussians to represent all modalities, which contradicts the distinct properties and granularities among modalities, resulting in suboptimal performance and redundant scene representations. In contrast, we disentangle Gaussians into shared and modality-specific components (\textit{i.e.}, multi-modal and single-modal Gaussians), consistently enhancing the representation capability for each modality and ensuring scalability to accommodate additional modalities.

\begin{figure*}[t]
    \centering
    \includegraphics[width=1.0\textwidth]{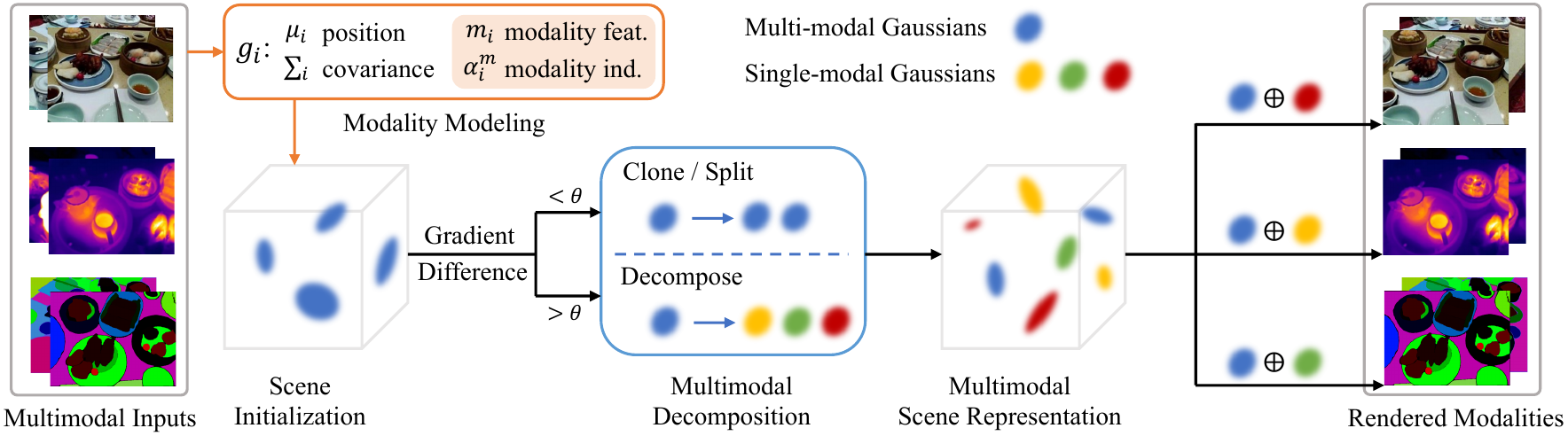}
    \caption{The General Framework of \textbf{MMOne}. Given multi-view multimodal inputs of the scene, we progressively construct a multimodal scene representation. Each modality is represented by our \textbf{modality modeling module}, which includes modality-specific features and a modality indicator. The densification process is integrated with our \textbf{multimodal decomposition mechanism}, which disentangles multimodal information based on gradient difference among modalities. Multimodal losses are combined to jointly optimize all modalities.}  
    \label{fig:framework}
    \vspace{-8pt}
\end{figure*}

%% file: sec/3_analysis.tex
\subsection{Preliminaries}

The vanilla Gaussian Splatting (3DGS) \cite{kerbl20233d} models a scene with a set of 3D Gaussians $\mathcal{G}=\{g_i|i=1,\cdots,N\}$. Each Gaussian $g_i$ is defined by its opacity $\alpha_i\in\mathbb{R}$, mean position $\mu_i\in\mathbb{R}^3$, covariance matrix $\Sigma_i\in\mathbb{R}^{3\times3}$ as $g_i(x)=e^{-\frac{1}{2}(x-\mu_i)^T\Sigma_i^{-1}(x-\mu_i)}$. Each Gaussian is also associated with color features $c_i\in\mathbb{R}^{d_c}$ (with $d_c=3$ color dimension), modeled by spherical harmonic (SH) coefficients.

To render other modalities such as thermal \cite{lu2025thermalgaussian} and language \cite{qin2024langsplat, zhou2024feature, wu2024opengaussian}, new feature vectors are introduced to model these modalities. The rendering process for these modalities is similar to RGB rendering. 3DGS first splats the Gaussians to the 2D image plane and then applies alpha-blending to compute the modality value for each pixel. This process can be generally formulated as:
\begin{equation}\label{eq:multimodality_render}
\begin{aligned}
    M(x) &= \sum_{i=1}^N T_i \cdot \alpha_i \cdot g_i^{2D}(x) \cdot m_i, \\
    \text{where} \quad T_i &= \prod_{j=1}^{i-1}(1-\alpha_j \cdot g_j^{2D}(x)),
\end{aligned}
\end{equation}
where $M$ represents a specific modality, $m_i$ denotes the modality feature vector (with $d_{m}$ feature dimension), and $g_i^{2D}(x)$ is the 2D projection of $g_i$ at pixel $x$. All attributes of the 3D Gaussians, including the modality-specific features, are learnable and optimized directly during training.

\subsection{Challenges in Multimodal Representation}

Modality conflicts arise when representing multiple modalities with a single scene representation. This is due to the intrinsic differences in properties and levels of granularity among different modalities. For example, as shown in \cref{fig:teaser}, the thermal modality is relatively coarse, the color modality is more detailed, and the language modality remains constant within objects or parts. Addressing these modality conflicts involves two main challenges.

\noindent\textbf{Property Disparity.} This refers to the inherent differences in the characteristics and attributes of data from various modalities. These disparities manifest in different forms, such as dimensionality and physical properties. For example, RGB representation requires three-dimensional features, whereas language representation necessitates a much higher dimensionality. Additionally, a piece of paper in front of a cup of hot tea would obscure the tea in the RGB and language modalities but not in the thermal modality.

\noindent\textbf{Granularity Disparity.} This denotes the inherent differences in the level of detail at which information is represented across various modalities. For example, the thermal modality is relatively coarse-grained, while the color modality is more fine-grained. Consequently, at the object boundaries, the thermal modality may favor fewer but larger Gaussians, whereas the color modality may require smaller but more numerous Gaussians. This challenge is exacerbated when different modalities share the same Gaussians.

%% file: sec/3_method.tex
\subsection{Overview}

We generally formulate the problem of representing multiple modalities in one scene as learning a scene representation to simultaneously represent multiple modalities, which can be readily extended to additional modalities.

Our framework, \textbf{MMOne}, as illustrated in \cref{fig:framework}, represents multiple modalities with one scene representation. Given the multi-view multimodal inputs of the scene, we progressively construct a multimodal scene representation capable of accommodating one or more modalities. Each modality is modeled by our \textbf{modality modeling module}, which includes modality-specific features and a modality indicator, as detailed in \cref{sec:modality_representation}. The densification process is integrated with our \textbf{multimodal decomposition mechanism}, which disentangles multimodal information based on modality differences, as discussed in \cref{sec:multimodal_decompostion}.

During training, each modality is rendered independently (as described in  \cref{eq:modality_indicator}), and the loss for each modality is computed separately. The overall loss is the sum of the individual losses for all modalities: $\mathcal{L}=\sum_{i=1}^m\mathcal{L}_{M_i}$, where $M_i$ represents a modality and $m$ denotes the number of modalities. The form of $\mathcal{L}_{M_i}$ varies due to the unique characteristics of each modality. All attributes of Gaussians, including modality-specific features and modality indicators, are optimized simultaneously through backpropagation. During inference, we can selectively render different modalities to accommodate various application scenarios.


\subsection{Representing Multiple Modalities}
\label{sec:modality_representation}

To address the property disparity, we propose a modality modeling module to model each modality, which can represent the distinct properties of different modalities. The module includes modality-specific features $m_i\in\mathbb{R}^{d_m}$ (with $d_m$ feature dimension) and a modality indicator $\alpha^m\in[0,1]$ (denoting the modality indicator of modality $M$). The rendering process for this modality can be formulated as:
\begin{equation}\label{eq:modality_indicator}
\begin{aligned}
    M(x) &= \sum_{i=1}^N T_i^m \cdot \alpha_i^m \cdot g_i^{2D}(x) \cdot m_i, \\
    \text{where} \quad T_i^m &= \prod_{j=1}^{i-1}(1-\alpha_j^m \cdot g_j^{2D}(x)).
\end{aligned}   
\end{equation}

Our proposed modality indicator is designed to capture the unique properties of each modality, rather than using a shared opacity across different modalities. In addition to the weighting mechanism described in \cref{eq:modality_indicator}, the modality indicator also functions as a ``\textbf{switch}'' for specific modalities, enabling selective deactivation during the rendering process. When certain modalities are deactivated (\textit{i.e.}, in an ``off'' state), the properties of the Gaussians (\textit{e.g.}, location, size, and orientation) are influenced solely by the remaining ``active'' modalities. This approach allows us to explicitly model the varying granularities of different modalities, as each modality requires a different number of Gaussians to represent. This forms the basis of our multimodal decomposition mechanism discussed in \cref{sec:multimodal_decompostion}.


The modality ``switch'' function is implemented by selectively skipping the rendering of certain modalities in the CUDA rasterization process, thereby freezing updates for those modalities. During training, our modality modeling module and other attributes of Gaussians are optimized simultaneously to capture modality-specific properties and distinguish information from different modalities.

\begin{figure}[t]
    \centering
    \includegraphics[width=1\linewidth]{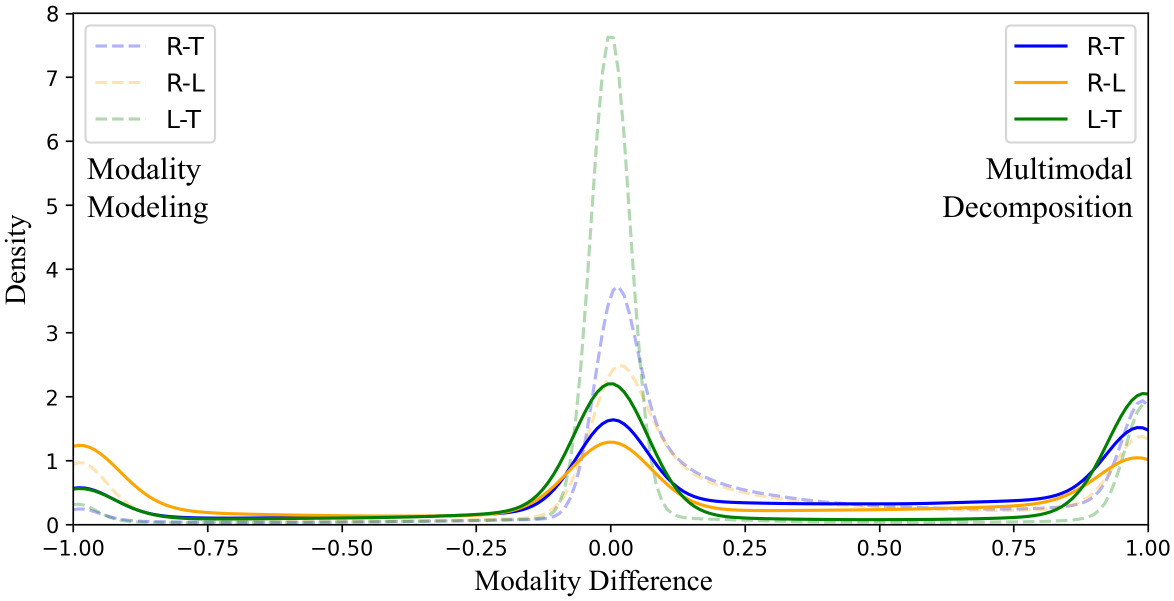}
    \caption{Distribution differences among different modality indicators using our modality modeling module (``dashed'' line) and multimodal decomposition mechanism (``solid'' line). ``R-T'' refers to $\alpha^R-\alpha^T$ in Gaussians, and so on. }   \label{fig:distribution_differencce}
    \vspace{-12pt}
\end{figure}

As illustrated in \cref{fig:distribution_differencce}, the dashed lines indicate the distinct distributions of modality indicators for different modalities. These differences enable the proposed modality modeling module to effectively capture and represent the unique properties of each modality.

\subsection{Multimodal Decomposition}
\label{sec:multimodal_decompostion}

To tackle the granularity disparity, we propose a simple yet effective multimodal decomposition mechanism to separate multi-modal Gaussians into single-modal Gaussians, accommodating the varying granularities of different modalities. This method disentangles multimodal information into shared and modality-specific components, leading to a more compact and efficient multimodal scene representation.

\noindent\textbf{Multimodal Prune.} 3DGS initializes Gaussians using a sparse point cloud from Structure-from-Motion (SfM) and employs an adaptive density control strategy to dynamically add and remove Gaussians. Gaussians are pruned when they exhibit low opacity or an overlarge screen space size. This approach is feasible when Gaussians are associated with one modality. However, conflicts arise when Gaussians are associated with multiple modality indicators representing different modalities. For example, if one modality indicator is low while another is high, simply pruning the Gaussian (\textit{i.e.}, ``\textbf{Hard Prune}'') may adversely affect other modalities.

\begin{figure}[t]
    \centering
    \includegraphics[width=1\linewidth]{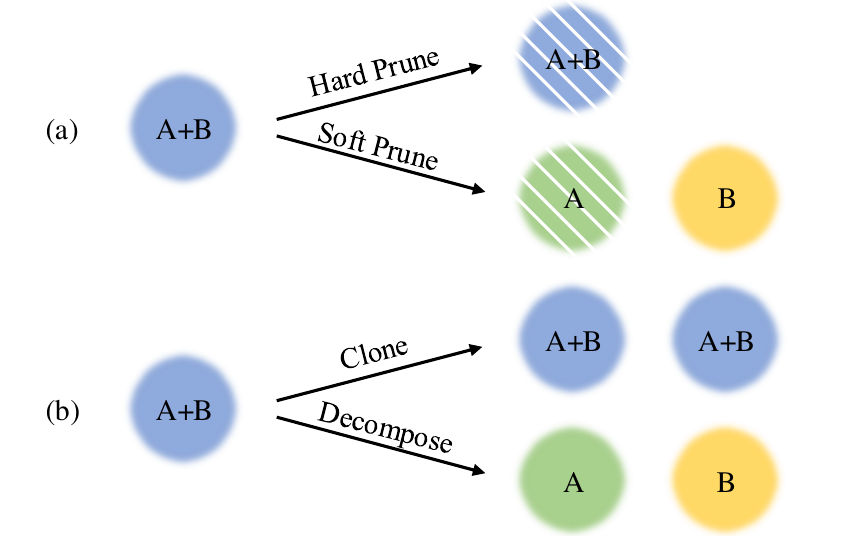}
    \caption{\textbf{Multimodal Decomposition Mechanism.} (a) Multimodal Prune: ``Hard Prune'' refers to directly pruning the Gaussian, while ``Soft Prune'' involves pruning modality ``A'' but retaining modality ``B''. (b) Multimodal Decomposition: Instead of cloning the Gaussian, multimodal decomposition disentangles a multi-modal Gaussian into multiple single-modal Gaussians.}   
    \label{fig:multimodal_decomposition}
    \vspace{-12pt}
\end{figure}

To address this, we propose ``\textbf{Soft Prune}'', which refers to pruning a specific modality rather than the entire Gaussian, as shown in \cref{fig:multimodal_decomposition}. This is achieved by setting the corresponding modality indicator to an ``off'' state, thereby excluding that modality from the rendering process.

However, simply ``soft pruning'' modalities can lead to redundant single-modal Gaussians, which may negatively impact the mutual improvement among multiple modalities. Therefore, we increase the threshold for pruning single-modal Gaussians to reduce the number of unimportant single-modal Gaussians, thereby encouraging the learning of the shared properties among different modalities.

\noindent\textbf{Multimodal Decomposition.} In the densification process of 3DGS, Gaussians with gradients exceeding a certain threshold are either cloned or split based on their scales. During the learning of multimodal representation, gradients from different modalities backpropagate to the same Gaussian. Conflicts arise when these gradients cancel each other out, leading to suboptimal results for all modalities.

To address this, we propose multimodal decomposition, which disentangles the gradients from different modalities and decomposes multimodality when the gradient difference exceeds a certain threshold. Specifically, we accumulate the gradients from different modalities (\textit{i.e.}, $g_{m_i}$ and $g_{m_j}$) and the gradient difference $gd_{ij}$ between modality $m_i$ and modality $m_j$ is formulated as:
\begin{equation}\label{eq:grad_diff}
\begin{aligned}
    gd_{ij} = norm(g_{m_i} - g_{m_j}).
\end{aligned}   
\end{equation}

We decompose the identified multi-modal Gaussians into multiple single-modal Gaussians, as shown in \cref{fig:multimodal_decomposition}. These decomposed Gaussians are then optimized separately based on their modality-specific losses. As illustrated in \cref{fig:distribution_differencce}, our multimodal decomposition mechanism preserves the distinct distributions of different modality indicators and removes redundant multi-modal Gaussians.

%% file: sec/4_experiments.tex
\begin{table*}[!htpb]
    \centering
    \fontsize{6}{4}\selectfont
    \caption{Quantitative evaluation of RGB-Thermal. ``R'' and ``T'' denote RGB and Thermal. ThermalGaussian \cite{lu2025thermalgaussian} is shortened as ``T-GS''. }   
    
    \resizebox{1.0\linewidth}{!}{
        \begin{tabular}{cccccccccccccc}
        \toprule[0.8pt]
        \multicolumn{1}{c}{M}
        &\multicolumn{1}{c}{Metric}
        &\multicolumn{1}{c}{Method}
        &\multicolumn{1}{c}{Dim}
        &\multicolumn{1}{c}{DS}
        &\multicolumn{1}{c}{Ebk}
        &\multicolumn{1}{c}{RB}
        &\multicolumn{1}{c}{Trk}
        &\multicolumn{1}{c}{RK}
        &\multicolumn{1}{c}{Bldg}
        &\multicolumn{1}{c}{II}
        &\multicolumn{1}{c}{Pt}
        &\multicolumn{1}{c}{LS}
        &\multicolumn{1}{c}{Avg.} \\
    
        \toprule[0.6pt]
        \multirow{3}*{}
        &\multirow{3}*{PSNR $\uparrow$ }
        &3DGS & 23.91 & 20.43 & 26.77 & 27.80 & 23.00 & 20.79 & 20.95 & 23.96 & 24.91 & 20.20 & 23.27 \\
        &&T-GS & \underline{24.38} & \underline{21.76} & \underline{26.85} & \underline{28.12} & \underline{23.66} & \underline{23.14} & \textbf{24.19} & \underline{24.55} & \underline{25.48} & \underline{21.71} & \underline{24.38} \\
        &&\textbf{MMOne} & \textbf{24.65} & \textbf{22.05} & \textbf{27.43} & \textbf{29.03} & \textbf{23.96} & \textbf{24.12} & \underline{24.16} & \textbf{25.65} & \textbf{26.01} & \textbf{21.81} & \textbf{24.89} \\
        \cmidrule{2-14}
        \multirow{3}*{R}
        &\multirow{3}*{SSIM $\uparrow$ }
        &3DGS & 0.847 & 0.748 & \underline{0.901} & 0.910 & 0.815 & 0.772 & 0.791 & 0.872 & \underline{0.859} & 0.696 & 0.821 \\
        &&T-GS & \underline{0.858} & \underline{0.797} & 0.900 & \textbf{0.920} & \underline{0.832} & \underline{0.822} & \textbf{0.849} & \underline{0.884} & 0.855 & \textbf{0.739} & \underline{0.846} \\
        &&\textbf{MMOne} & \textbf{0.862} & \textbf{0.810} & \textbf{0.918} & \underline{0.916} & \textbf{0.845} & \textbf{0.842} & \underline{0.847} & \textbf{0.897} & \textbf{0.876} & \underline{0.727} & \textbf{0.854} \\
        \cmidrule{2-14}
        \multirow{3}*{}
        &\multirow{3}*{LPIPS $\downarrow$ }
        &3DGS & \textbf{0.194} & 0.299 & 0.171 & \underline{0.201} & 0.238 & 0.217 & 0.228 & 0.188 & \underline{0.183} & \underline{0.280} & 0.220 \\
        &&T-GS & \textbf{0.194} & \textbf{0.253} & \underline{0.169} & \textbf{0.199} & \textbf{0.224} & \underline{0.184} & \textbf{0.170} & \underline{0.186} & 0.195 & \textbf{0.268} & \textbf{0.204} \\
        &&\textbf{MMOne} & \underline{0.203} & \underline{0.254} & \textbf{0.160} & 0.235 & \underline{0.226} & \textbf{0.178} & \underline{0.184} & \textbf{0.183} & \textbf{0.178} & 0.291 & \underline{0.209} \\

        \toprule[0.6pt]
        \multirow{3}*{}
        &\multirow{3}*{PSNR $\uparrow$ }
        &3DGS & 26.21 & 20.28 & 20.78 & 26.46 & 23.93 & \underline{27.17} & 25.39 & \underline{29.90} & 22.33 & 18.68 & 24.11 \\
        &&T-GS & \underline{26.46} & \textbf{22.28} & \underline{23.31} & \underline{27.17} & \underline{24.57} & 26.33 & \underline{26.72} & 29.86 & \textbf{26.16} & \underline{22.27} & \underline{25.51} \\
        &&\textbf{MMOne} & \textbf{26.90} & \underline{21.81} & \textbf{23.79} & \textbf{27.39} & \textbf{25.44} & \textbf{27.65} & \textbf{27.06} & \textbf{30.27} & \underline{26.05} & \textbf{22.52} & \textbf{25.89} \\
        \cmidrule{2-14}
        \multirow{3}*{T}
        &\multirow{3}*{SSIM $ \uparrow$}
        &3DGS & \underline{0.890} & 0.816 & 0.814 & 0.914 & \underline{0.853} & \underline{0.928} & 0.873 & \underline{0.897} & 0.842 & 0.760 & 0.859 \\
        &&T-GS & 0.886 & \underline{0.835} & \underline{0.862} & \underline{0.919} & 0.849 & 0.922 & \underline{0.888} & 0.896 & \underline{0.883} & \underline{0.850} & \underline{0.879} \\
        &&\textbf{MMOne} & \textbf{0.894} & \textbf{0.840} & \textbf{0.874} & \textbf{0.926} & \textbf{0.870} & \textbf{0.933} & \textbf{0.902} & \textbf{0.906} & \textbf{0.895} & \textbf{0.861} & \textbf{0.890} \\
        \cmidrule{2-14}
        \multirow{3}*{}
        &\multirow{3}*{LPIPS $\downarrow$ }
        &3DGS & \underline{0.126} & 0.240 & 0.314 & \underline{0.213} & 0.160 & \underline{0.126} & 0.223 & \underline{0.088} & 0.265 & 0.383 & 0.214 \\
        &&T-GS & 0.129 & \underline{0.210} & \underline{0.203} & \textbf{0.198} & \underline{0.155} & \textbf{0.124} & \textbf{0.177} & 0.091 & \textbf{0.181} & \textbf{0.248} & \textbf{0.172} \\
        &&\textbf{MMOne} & \textbf{0.125} & \textbf{0.194} & \textbf{0.201} & \underline{0.213} & \textbf{0.142} & 0.127 & \underline{0.198} & \textbf{0.083} & \underline{0.205} & \underline{0.272} & \underline{0.176} \\
        \bottomrule[0.8pt]
        \end{tabular}
    }
    \vspace{-8pt}
\label{table:thermal_quantitative}
\end{table*}

To evaluate our approach, we select three representative modalities for evaluation: RGB, thermal, and language. We conducted experiments on RGB-Thermal, RGB-Language, and RGB-Thermal-Language combinations to demonstrate the effectiveness of the proposed method. 


\noindent\textbf{Datasets.} We conduct experimental evaluation on two real-world datasets. 1) \textbf{RGBT-Scenes dataset} \cite{lu2025thermalgaussian}, collected using a handheld thermal-infrared camera FLIR E6 PRO, comprises over 1000 aligned RGB and thermal images from 10 diverse scenes, including both indoor and outdoor environments with various object sizes and temperature variations. 2) \textbf{LERF dataset} \cite{kerr2023lerf}, captured using the Polycam application of an iPhone, includes complex in-the-wild scenes. We utilize the extended version from LangSplat \cite{qin2024langsplat}, which provides ground truth masks for text queries and additional challenging localization samples.

\noindent\textbf{Metrics.} We evaluate the RGB and thermal modalities using standard image quality metrics: Peak Signal-to-Noise Ratio (PSNR), Structural Similarity Index (SSIM), and Learned Perceptual Image Patch Similarity (LPIPS). These metrics assess the quality of the reconstructed images from novel views. For the language modality, we report the mean Intersection over Union (mIoU) and localization accuracy (\%) for open-vocabulary semantic segmentation and object localization following LangSplat \cite{qin2024langsplat}.

\noindent\textbf{Implementation Details.} Our method is implemented based on the 3DGS framework \cite{kerbl20233d} using PyTorch. All experimental settings and default parameters are consistent with 3DGS. We set the threshold for multimodal decomposition to 0.0002. We modify the rasterization module of 3DGS to render thermal and language modalities similar to \cite{lu2025thermalgaussian, qin2024langsplat}. Ground-truth semantic features are obtained following LangSplat \cite{qin2024langsplat} using the SAM ViT-H model \cite{kirillov2023segment} and the OpenCLIP ViT-B/16 model \cite{radford2021learning}. We train each comparative experiment for 30K iterations. All experiments are conducted on a single NVIDIA 4090 GPU.

\subsection{Evaluation on RGB-Thermal}

We compare our method with two baselines: 1) 3DGS \cite{kerbl20233d}, which trains the original 3DGS using RGB and thermal images separately, and 2) ThermalGaussian \cite{lu2025thermalgaussian}, which trains RGB and thermal jointly with a shared opacity. For a fair comparison, we adopt the same experimental settings and loss functions (\textit{e.g.}, smoothness loss on thermal images) as used in \cite{lu2025thermalgaussian}. All comparative experiments utilize the same sparse point cloud and camera poses obtained from COLMAP \cite{schonberger2016structure} results on RGB images.

The quantitative results of various methods are presented in Table \ref{table:thermal_quantitative}. Note that we reproduce results on the ``Truck'' scene to align with the train/test split of other scenes and 3DGS. The table demonstrates that our method consistently outperforms the strong baseline, ThermalGaussian \cite{lu2025thermalgaussian}, across most scenarios. Despite approaching performance saturation, our approach shows an average PSNR improvement of 0.5dB in RGB rendering and 0.4dB in thermal rendering, highlighting the effectiveness of our method to capture modality-specific properties and granularities. Notably, our method employs one-third of the Gaussians used by ThermalGaussian, yet achieves superior performance.

\begin{figure*}[ht!]
    \centering
    \includegraphics[width=1\linewidth]{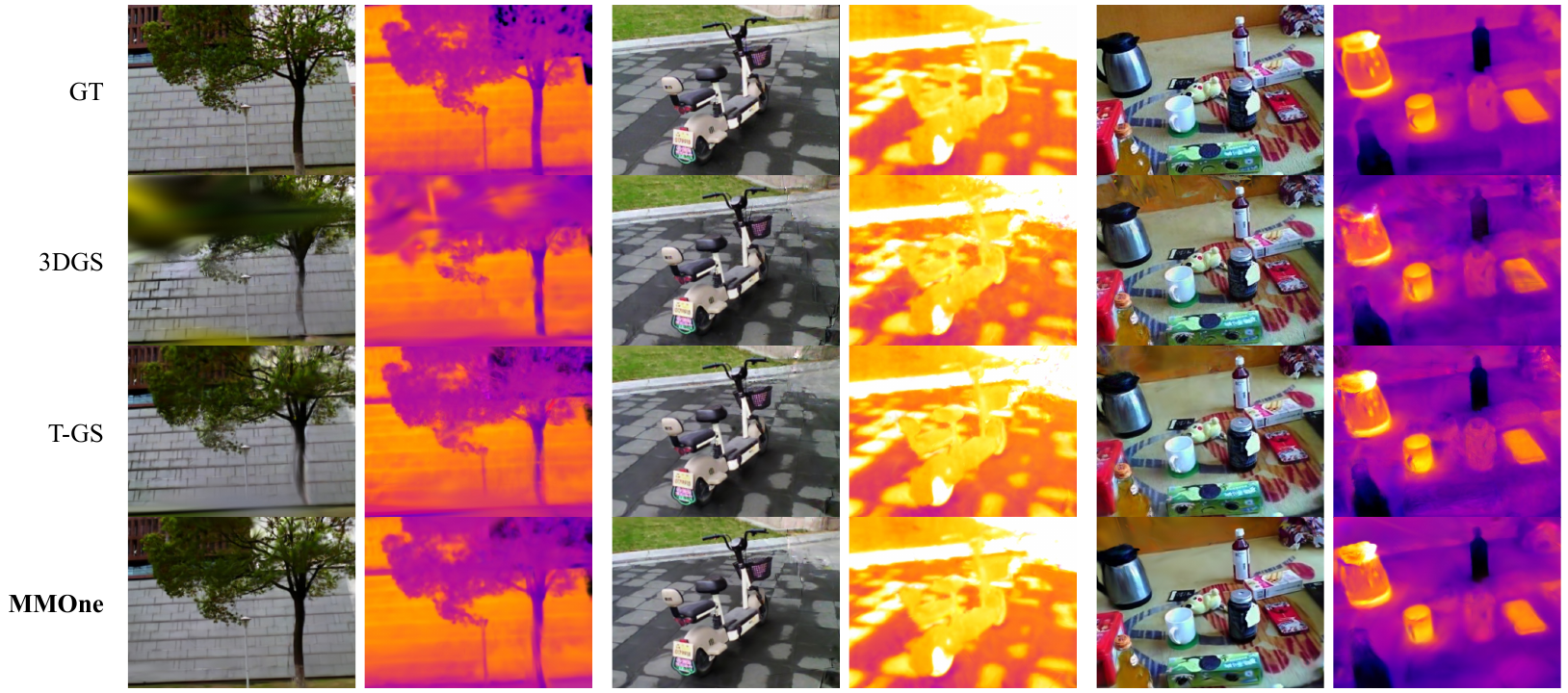}
    \caption{Qualitative results of RGB-Thermal among      our MMOne, 3DGS, and ThermalGaussian (abbreviated as ``T-GS'').}  
    \label{fig:thermal_qualitative}
    \vspace{-4pt}
\end{figure*}

The rendered RGB and thermal images are presented in \cref{fig:thermal_qualitative}. These qualitative results align with the quantitative findings in \cref{table:thermal_quantitative}. Our modality modeling module maintains the sharpness of the RGB modality and the smoothness of the thermal modality, resulting in clearer colors and more defined boundaries. Furthermore, our multimodal decomposition mechanism mitigates modality conflicts, enhancing the consistency of both modalities with reduced noise.

\begin{table}[htpb!]
    \centering
    \Huge
    \caption{Quantitative evaluation of RGB-Language. ``R'' and ``L'' denote RGB and Language. LangSplat \cite{qin2024langsplat} is shortened as ``LS*'' and ``LS-J'' denotes the joint training baseline based on LangSplat. }
    \resizebox{1.0\linewidth}{!}{
    \begin{tabular}{cccccccc}
    \toprule[3pt]
    \multicolumn{1}{c}{M}
    &\multicolumn{1}{c}{Metric}
    &\multicolumn{1}{c}{Method}
    &\multicolumn{1}{c}{Figurines}
    &\multicolumn{1}{c}{Ramen}
    &\multicolumn{1}{c}{Teatime}
    &\multicolumn{1}{c}{Kitchen}
    &\multicolumn{1}{c}{Avg.} \\

    \toprule[2pt]
    \multirow{3}*{}
    &\multirow{3}*{PSNR $\uparrow$ }
    &LS* & \textbf{24.31} & \underline{24.45} & \underline{23.79} & \underline{23.52} & \underline{24.02} \\
    & &LS-J & 22.43 & 24.29 & 23.14 & 23.06 & 23.23 \\
    & &\textbf{MMOne} & \underline{24.07} & \textbf{24.71} & \textbf{24.03} & \textbf{24.58} & \textbf{24.35} \\

    \cmidrule{2-8}
    \multirow{3}*{R}
    &\multirow{3}*{SSIM $\uparrow$ } 
    &LS* & \textbf{0.844} & \textbf{0.858} & \textbf{0.821} & 0.893 & \textbf{0.854} \\
    & &LS-J & 0.818 & 0.849 & 0.802 & 0.880 & 0.837 \\
    & &\textbf{MMOne} & \underline{0.837} & \underline{0.854} & \underline{0.817} & \textbf{0.896} & \underline{0.851} \\

    \cmidrule{2-8}
    \multirow{3}*{}
    &\multirow{3}*{LPIPS $\downarrow$} 
    &LS* & \textbf{0.215} & \textbf{0.194} & \textbf{0.273} & \textbf{0.196} & \textbf{0.220} \\
    & &LS-J & 0.259 & 0.222 & 0.320 & 0.225 & 0.257 \\
    & &\textbf{MMOne} & \underline{0.242} & \underline{0.222} & \underline{0.307} & \underline{0.203} & \underline{0.244} \\

    \toprule[2pt]
    \multirow{6.5}*{L}
    &\multirow{3}*{mIoU $\uparrow$ } 
    &LS* & 45.2 & 46.1 & 53.0 & 46.2 & 47.6 \\
    & &LS-J & \underline{58.6} & \underline{47.8} & \underline{60.0} & \underline{54.6} & \underline{55.3} \\
    & &\textbf{MMOne} & \textbf{58.9} & \textbf{48.0} & \textbf{62.0} & \textbf{57.5} & \textbf{56.6} \\

    \cmidrule{2-8}
    &\multirow{3}*{acc $\uparrow$} 
    &LS* & \underline{71.4} & \underline{60.6} & \underline{84.8} & 72.7 & 72.4 \\
    & &LS-J & \underline{71.4} & \underline{60.6} & \underline{84.8} & \textbf{77.3} & \underline{73.5} \\
        & &\textbf{MMOne} & \textbf{80.4} & \textbf{62.0} & \textbf{86.4} & \textbf{77.3} & \textbf{76.5} \\
    \bottomrule[3pt]
  \end{tabular}
  }
\label{table:lerf_quantitative}
\end{table}

\subsection{Evaluation on RGB-Language}

We evaluate our method against 2 baselines: 1) LangSplat \cite{qin2024langsplat}, which first reconstructs the scene geometry using RGB and then registers language features to the Gaussians, and 2) ``LS-J'', a RGB-language joint training baseline modified from LangSplat, which trains RGB and Language jointly using a shared opacity. Note that we reproduce LangSplat \cite{qin2024langsplat} after splitting the train/test set to evaluate both the RGB rendering and open-vocabulary queries. 


As shown in \cref{table:lerf_quantitative}, LangSplat \cite{qin2024langsplat} relies on the original 3DGS for RGB reconstruction, limiting its language understanding to this fixed RGB geometry. ``LS-J'' demonstrates strong performance in open-vocabulary localization and semantic segmentation, surpassing LangSplat by 7.7 in mIoU. However, due to the modality conflicts introduced by the language modality, its RGB results decline across all scenes. In contrast, our method excels in open-vocabulary queries while maintaining high RGB rendering performance. Notably, our PSNR even exceeds LangSplat, highlighting the mutual enhancements between modalities and the effectiveness of disentangling multimodality.

\begin{figure*}[htpb!]
    \centering
    \includegraphics[width=1.0\textwidth]{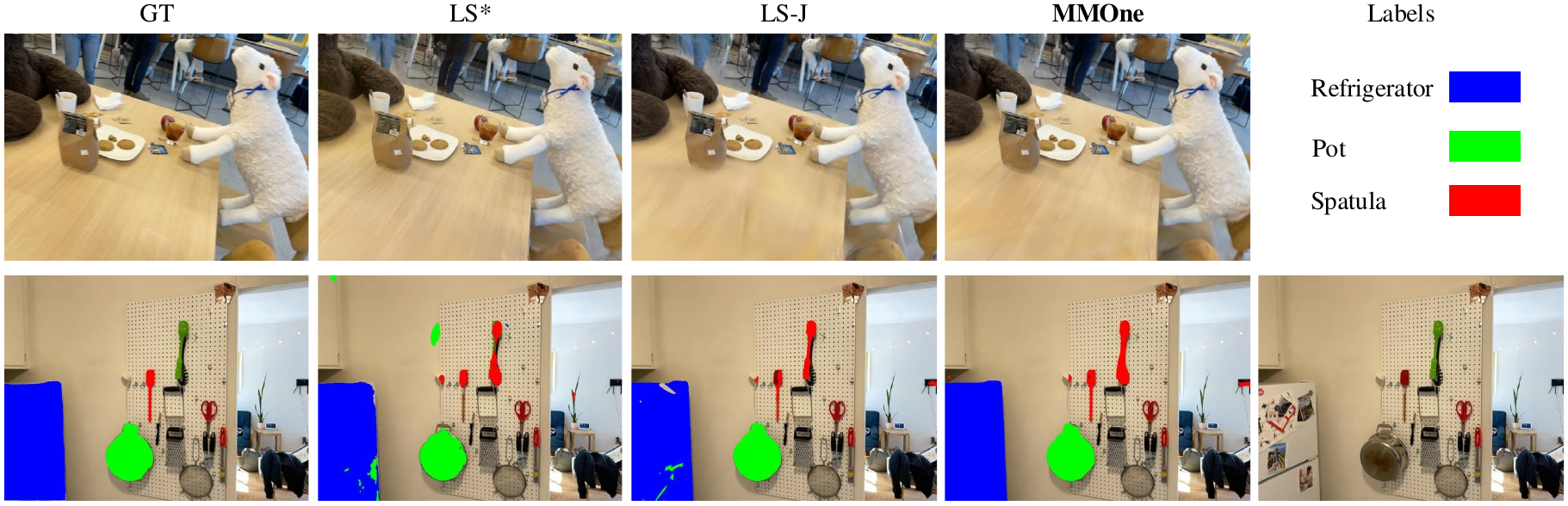}
    \caption{Qualitative results of RGB-Language among our MMOne, LangSplat (``LS*''), and joint training version of LangSplat (``LS-J''). The first row presents RGB renderings from novel views and the second row shows the open-vocabulary semantic segmentation results.   }  
    \label{fig:language_qualitative}
    \vspace{-14pt}
\end{figure*}

Both RGB and Language qualitative results are presented in \cref{fig:language_qualitative}. Our method preserves the high RGB rendering quality of 3DGS, with sharper object boundaries due to the integration of the language modality. With our modality modeling module and multimodal decomposition mechanism, the semantic masks generated from text queries are more accurately aligned with object surfaces and exhibit greater consistency within objects.

\begin{table}[htpb!]
    \centering
    \large
    \caption{Quantitative evaluation of RGB-Thermal-Language. ``R'', ``T'' and ``L'' denote RGB, Thermal and Language, respectively. ``MM-J'' represents our joint training baseline.}  
    \resizebox{1.0\linewidth}{!}{
    \begin{tabular}{cccccccc}
    \toprule[1.5pt]
    \multicolumn{1}{c}{M}
    &\multicolumn{1}{c}{Metric}
    &\multicolumn{1}{c}{Method}
    &\multicolumn{1}{c}{Dimsum}
    &\multicolumn{1}{c}{DS}
    &\multicolumn{1}{c}{LS}
    &\multicolumn{1}{c}{Truck}
    &\multicolumn{1}{c}{Avg.} \\

    \toprule[1pt]
    \multirow{2}*{}
    &\multirow{2}*{PSNR $\uparrow$ }
    &MM-J & 23.99 & 20.99 & 20.98 & 23.32 & 22.32 \\
    & &\textbf{MMOne} & \textbf{24.74} & \textbf{22.15} & \textbf{21.85} & \textbf{24.01} & \textbf{23.19} \\

    \cmidrule{2-8}
    \multirow{2}*{R}
    &\multirow{2}*{SSIM $\uparrow$ } 
    &MM-J & 0.854 & 0.782 & 0.721 & 0.827 & 0.796 \\
    & &\textbf{MMOne} & \textbf{0.863} & \textbf{0.814} & \textbf{0.723} & \textbf{0.846} & \textbf{0.812} \\

    \cmidrule{2-8}
    \multirow{2}*{}
    &\multirow{2}*{LPIPS $\downarrow$} 
    &MM-J & \textbf{0.199} & 0.277 & \textbf{0.281} & 0.232 & 0.247 \\
    & &\textbf{MMOne} & 0.204 & \textbf{0.251} & 0.296 & \textbf{0.228} & \textbf{0.245} \\

    \toprule[1pt]
    \multirow{2}*{}
    &\multirow{2}*{PSNR $\uparrow$ }
    &MM-J & 26.18 & 21.55 & 21.65 & 24.13 & 23.38 \\
    & &\textbf{MMOne} & \textbf{26.82} & \textbf{22.11} & \textbf{22.57} & \textbf{25.46} & \textbf{24.24} \\

    \cmidrule{2-8}
    \multirow{2}*{T}
    &\multirow{2}*{SSIM $\uparrow$ } 
    &MM-J & 0.886 & 0.828 & 0.840 & 0.842 & 0.849 \\
    & &\textbf{MMOne} & \textbf{0.893} & \textbf{0.848} & \textbf{0.860} & \textbf{0.868} & \textbf{0.867} \\

    \cmidrule{2-8}
    \multirow{2}*{}
    &\multirow{2}*{LPIPS $\downarrow$} 
    &MM-J & \textbf{0.130} & 0.227 & \textbf{0.274} & 0.168 & 0.200 \\
    & &\textbf{MMOne} & 0.131 & \textbf{0.190} & 0.279 & \textbf{0.147} & \textbf{0.187} \\

    \toprule[1pt]
    \multirow{2}*{L}
    &\multirow{2}*{mIoU $\uparrow$ } 
    &MM-J & 56.0 & 26.9 & 44.9 & 52.4 & 45.1 \\
    & &\textbf{MMOne} & \textbf{61.1} & \textbf{30.6} & \textbf{46.1} & \textbf{54.7} & \textbf{48.1} \\
    \bottomrule[1.5pt]
  \end{tabular}
  }
  \vspace{-4pt}
\label{table:thermal_language_quantitative}
\end{table}

\begin{table}[htpb!]
    \caption{Quantitative analysis of modality conflicts. ``R/T'' refers to training with RGB and thermal, while ``R/T/L'' indicates training with RGB, thermal, and language. The bolded values highlight the changes in RGB and thermal with the introduction of language. } 
    \centering
    \Huge
    \resizebox{1.0\linewidth}{!}{
    \begin{tabular}{lccccc}  
        \toprule[3pt]
        \multirow{2}{1.7cm}{Methods}
        &\multicolumn{2}c{RGB} 
        &\multicolumn{2}c{Thermal} \\
  
        \cmidrule(lr){2-3} \cmidrule(lr){4-5}
        &\multicolumn{1}{c}{PSNR}
        &\multicolumn{1}{c}{SSIM}
        &\multicolumn{1}{c}{PSNR}
        &\multicolumn{1}{c}{SSIM} \\

        \toprule[2pt]
        T-GS (R/T) & 22.88 & 0.807 & 23.90 & 0.855 \\
        \textbf{MMOne} (R/T) & 23.12 & 0.811 & 24.17 & 0.866 \\

        \toprule[2pt]

        MM-J (R/T/L) & 22.32 \textbf{(-0.56)} & 0.796 \textbf{(-0.011)} & 23.38 \textbf{(-0.52)} & 0.849 \textbf{(-0.006)} \\
        \textbf{MMOne} (R/T/L) & 23.19 \textbf{(+0.07)} & 0.812 \textbf{(+0.001)} & 24.24 \textbf{(+0.07)} & 0.867 \textbf{(+0.001)} \\

    \bottomrule[3pt]
    \end{tabular}
    }
    \vspace{-4pt}
\label{table:modality_conflicts}
\end{table}

\subsection{Evaluation on RGB-Thermal-Language}

To evaluate our method on more than two modalities, we conduct experiments to learn the RGB-thermal-language representation. Since no existing RGB-thermal-language dataset is available, we select four scenes from the RGBT-Scenes dataset \cite{lu2025thermalgaussian}, which include multiple objects with aligned RGB and thermal images. We manually annotate the ground-truth semantic masks for open-vocabulary queries following LangSplat \cite{qin2024langsplat}. Specifically, we obtain ground-truth language features with CLIP \cite{radford2021learning} and SAM \cite{kirillov2023segment}. We only employ the medium level of SAM masks for the language modality learning as they best align with object-level granularity. Given the absence of established methods for representing RGB-thermal-language scenes, we compare our method with a self-implemented RGB-thermal-language joint training baseline that trains RGB, thermal, and language jointly using a shared opacity. Our experimental settings and loss functions are consistent with \cite{lu2025thermalgaussian, qin2024langsplat} without additional modifications.

\begin{figure}[htpb!]
    \centering
    \vspace{-2pt}
    \includegraphics[width=1\linewidth]{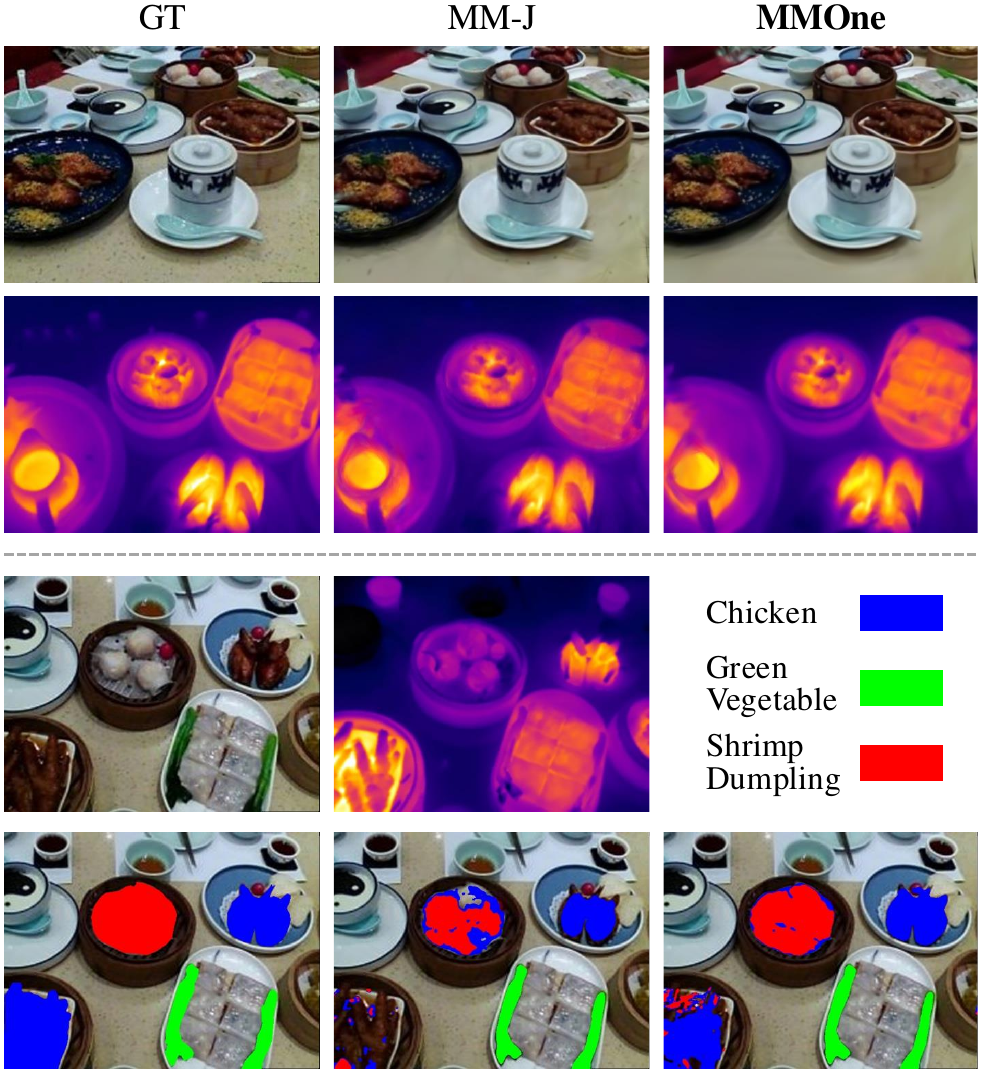}
    \caption{Qualitative results of RGB-Thermal-Language between our MMOne and the joint training baseline (``MM-J''). The first two rows present RGB and thermal renderings from novel views, the third row shows the reference images for text queries, and the fourth row displays the open-vocabulary query results.  }  
    \label{fig:thermal_language_qualitative}
    \vspace{-10pt}
\end{figure}

The quantitative results are presented in \cref{table:thermal_language_quantitative}. Our method achieves superior results across all three modalities, indicating the effectiveness of our modality modeling module in modeling modality-specific properties and our multimodal decomposition mechanism that disentangle multimodality to mitigate modality conflicts. As shown in \cref{fig:thermal_language_qualitative}, the qualitative results demonstrate that our method consistently improves the rendering quality of RGB and thermal. Additionally, the masks generated from text queries are better aligned with object surfaces.

Notably, as shown in \cref{table:modality_conflicts}, the RGB and thermal rendering quality significantly degrades when incorporating language into the joint training baselines (i.e., ``T-GS'' and ``MM-J''), underscoring the impact of modality conflicts as the number of modalities increases. In contrast, our method not only surpasses baselines with a large margin, but also slightly enhances RGB and thermal rendering quality with the language modality. This demonstrates the effectiveness of our method to handle multiple modalities and scalability to accommodate more modalities by decomposing multimodality into shared and modality-specific components.

\begin{table}[htpb!]
    \caption{\textbf{Ablation Studies.} We conduct ablation studies on RGB-Thermal-Language by gradually adding components to our joint training baseline, ``MM-J''. ``MM'' refers to our modality modeling module. ``H'' and ``S'' denote ``Hard'' and ``Soft'', respectively. } 
    \centering
    \huge
    \resizebox{1.0\linewidth}{!}{
    \begin{tabular}{lcccccccc}  
        \toprule[2.5pt]
        \multirow{2}{1.7cm}{Methods}
        &\multicolumn{3}c{RGB} 
        &\multicolumn{3}c{Thermal}
        &\multicolumn{1}c{Lang}
        &\multirow{2}{1.5cm}{Num\\$\times 10^4$} \\
  
        \cmidrule(lr){2-4} \cmidrule(lr){5-7} \cmidrule(lr){8-8}
        &\multicolumn{1}{c}{PSNR}
        &\multicolumn{1}{c}{SSIM}
        &\multicolumn{1}{c}{LPIPS}
        &\multicolumn{1}{c}{PSNR}
        &\multicolumn{1}{c}{SSIM}
        &\multicolumn{1}{c}{LPIPS}
        &\multicolumn{1}{c}{mIoU} \\

        \toprule[1pt]
        MM-J & 22.32 & 0.796 & 0.247 & 23.38 & 0.849 & 0.200 & 45.1 & 32.9 \\

        \toprule[1pt]
        + MM & 22.38 & 0.797 & \textbf{0.240} & 23.73 & 0.856 & 0.208 & 45.3 & 29.0 \\

        \toprule[1pt]
        Prune (H) & 22.67 & 0.801 & 0.247 & 23.86 & 0.863 & \textbf{0.184} & 46.9 & 13.4 \\
        Prune (S) & \underline{22.98} & \underline{0.808} & 0.248 & \underline{23.99} & \underline{0.865} & 0.188 & \underline{47.0} & \underline{10.6}  \\

        \toprule[1pt]
        Decomp. & \textbf{23.19} & \textbf{0.812} & \underline{0.245} & \textbf{24.24} & \textbf{0.867} & \underline{0.187} & \textbf{48.1} & \textbf{9.9} \\

    \bottomrule[2.5pt]
    \end{tabular}
    }
    \vspace{-14pt}
\label{table:ablation}
\end{table}

\subsection{Ablation Studies}

To evaluate the effectiveness of each component, we conduct ablation studies on representing RGB-thermal-language scenes, as modality conflicts are amplified when the number of modalities increases. As shown in \cref{table:ablation}, we begin by evaluating the impact of our modality modeling module. With the proposed modality indicator, this module enhances the capability to represent modality-specific properties, resulting in consistent improvements across the RGB, thermal and language modalities.

Next, we evaluate the impact of multimodal prune, comparing ``Hard Prune'' and ``Soft Prune''. Our proposed ``Soft Prune'' achieves better results across all modalities, demonstrating its effectiveness in mitigating the adverse effects on other modalities during pruning. It also significantly reduces the number of Gaussians, leading to a more compact scene representation. Finally, by incorporating our multimodal decomposition mechanism, which disentangles multimodality based on the gradient differences among modalities, we achieve exceptional results across all metrics. This indicates that employing different numbers of Gaussians to represent each modality can enhance overall performance, aligning with the varying levels of granularity of different modalities. Notably, our method results in a compact and efficient multimodal scene representation, using less than one-third of the number of Gaussians of the baseline while achieving superior performance.

%% file: sec/5_conclusion.tex
In this work, we propose MMOne, a general framework for representing multiple modalities in one scene,  which can be readily extended to additional modalities. Our method includes a modality modeling module and a multimodal decomposition mechanism, designed to capture modality-specific properties of different modalities and disentangle multimodal information into shared and modality-specific components. Extensive experiments across various modality combinations demonstrate the effectiveness and scalability of our method. 
Future work will focus on incorporating camera poses into the learning process  and extending multimodal scene representation to dynamic scenes.

%% file: sec/6_appendix.tex
\maketitlesupplementary

\section{Additional Implementation Details}

In this section, we introduce the implementation details of the RGB-Thermal, RGB-Language, and RGB-Thermal-Language experiments. We also introduce the hyperparameter settings in our proposed module. 

\noindent\textbf{RGB-Thermal.} We adhere to the experimental settings of ThermalGaussian for a fair comparison. Specifically, we use spherical harmonic coefficients to model the thermal modality and adjust our thermal rasterization to render thermal images akin to RGB images. The loss function mirrors that of, incorporating a smoothness loss for the thermal modality with $\lambda_{smooth}$ set to 0.6. The weights for both RGB and thermal losses are set to 0.5.

\noindent\textbf{RGB-Language.} Following LangSplat, we modify our language rasterization to render three-dimensional language features. The language loss is defined as the L1 loss between the ground-truth and rendered feature maps. The weights for both RGB and language losses are set to 0.5. For open-vocabulary localization and semantic segmentation, we adopt the same procedure as LangSplat. 

\noindent\textbf{RGB-Thermal-Language.} We employ the same thermal and language rasterization process as used in the two-modality evaluations. The same thermal and language losses are applied. The weights for RGB and thermal losses are set to 0.5, and the weight for language loss is set to 0.2.

\noindent\textbf{Hyperparameters.} In our proposed ``Soft Prune'', we set the pruning threshold for single-modal Gaussians to 0.5, effectively removing unimportant Gaussians and resulting in a more compact scene representation. For our multimodal decomposition mechanism, we employ the L2 norm to calculate gradient differences among modalities. This decomposition is integrated into the densification process. If the gradient difference between two modalities exceeds 0.0002, we decompose the multi-modal Gaussian into multiple single-modal Gaussians.

\section{Additional Ablation Studies}

To further investigate the sensitivity of the threshold setting of multimodal decomposition, we conduct an additional ablation study. As shown in \cref{table:ablation_threshold}, our chosen threshold consistently achieves superior performance across all metrics. Moreover, we observe only a slight performance drop when the threshold is adjusted, highlighting the robustness of our proposed method.

\begin{table}[htpb!]
    \caption{ Ablation for the threshold of multimodal decomposition. } 
    \centering
    \Huge
    \resizebox{1.0\linewidth}{!}{
    \begin{tabular}{lcccccccc}  
        \toprule[3pt]
        \multirow{2}{3.5cm}{Threshold}
        &\multicolumn{3}c{RGB} 
        &\multicolumn{3}c{Thermal}
        &\multicolumn{1}c{Lang}
        &\multirow{2}{1.7cm}{Num\\$\times 10^4$} \\
  
        \cmidrule(lr){2-4} \cmidrule(lr){5-7} \cmidrule(lr){8-8}
        &\multicolumn{1}{c}{PSNR $\uparrow$}
        &\multicolumn{1}{c}{SSIM $\uparrow$}
        &\multicolumn{1}{c}{LPIPS $\downarrow$}
        &\multicolumn{1}{c}{PSNR $\uparrow$}
        &\multicolumn{1}{c}{SSIM $\uparrow$}
        &\multicolumn{1}{c}{LPIPS $\downarrow$}
        &\multicolumn{1}{c}{mIoU $\uparrow$} \\

        \toprule[2pt]
        0.0001 & 23.11 & 0.809 & 0.246 & 23.94 & 0.864 & 0.190 & 47.1 & \textbf{9.5} \\
        
        \textbf{0.0002} & \textbf{23.19} & \textbf{0.812} & \textbf{0.245} & \textbf{24.24} & \textbf{0.867} & \textbf{0.187} & \textbf{48.1} & 9.9 \\

        0.0003 & 23.13 & 0.810 & 0.246 & 24.04 & 0.865 & \textbf{0.187} & 47.3 & 10.1 \\
        0.0004 & 23.03 & 0.808 & 0.246 & 24.15 & 0.866 & \textbf{0.187} & 47.0 & 10.3 \\

    \bottomrule[3pt]
    \end{tabular}
    }
\label{table:ablation_threshold}
\end{table}

\begin{table}[htpb!]
    \centering
    \large
    \caption{Full ablation studies on RGB-Thermal-Language by gradually adding components to our joint training baseline ``MM-J''. ``MM'' refers to our modality modeling module. ``H'' and ``S'' denote ``Hard'' and ``Soft'', respectively.}  
    \resizebox{1.0\linewidth}{!}{
    \begin{tabular}{cclccccc}
    \toprule[1.5pt]
    \multicolumn{1}{c}{M}
    &\multicolumn{1}{c}{Metric}
    &\multicolumn{1}{l}{Method}
    &\multicolumn{1}{c}{Dimsum}
    &\multicolumn{1}{c}{DS}
    &\multicolumn{1}{c}{LS}
    &\multicolumn{1}{c}{Truck}
    &\multicolumn{1}{c}{Avg.} \\

    \toprule[1pt]
    \multirow{5}*{}
    &\multirow{5}*{PSNR $\uparrow$ }
    &MM-J & 23.99 & 20.99 & 20.98 & 23.32 & 22.32 \\
    & &+ MM & 24.16 & 21.41 & 20.92 & 23.02 & 22.38 \\
    & &Prune (H) & 24.14 & 21.45 & \underline{21.80} & 23.27 & 22.67 \\
    & &Prune (S) & \underline{24.69} & \underline{21.99} & 21.78 & \underline{23.45} & \underline{22.98} \\
    & &Decomp. & \textbf{24.74} & \textbf{22.15} & \textbf{21.85} & \textbf{24.01} & \textbf{23.19} \\

    \cmidrule{2-8}
    \multirow{5}*{R}
    &\multirow{5}*{SSIM $\uparrow$ } 
    &MM-J & 0.854 & 0.782 & \underline{0.721} & 0.827 & 0.796 \\
    & &+ MM & 0.856 & 0.795 & 0.716 & 0.820 & 0.797 \\
    & &Prune (H) & 0.855 & 0.801 & 0.718 & 0.831 & 0.801 \\
    & &Prune (S) & \textbf{0.864} & \underline{0.812} & 0.720 & \underline{0.835} & \underline{0.808} \\
    & &Decomp. & \underline{0.863} & \textbf{0.814} & \textbf{0.723} & \textbf{0.846} & \textbf{0.812} \\

    \cmidrule{2-8}
    \multirow{5}*{}
    &\multirow{5}*{LPIPS $\downarrow$} 
    &MM-J & \underline{0.199} & 0.277 & \underline{0.281} & \underline{0.232} & 0.247 \\
    & &+ MM & \textbf{0.196} & 0.255 & \textbf{0.274} & 0.236 & \textbf{0.240} \\
    & &Prune (H) & 0.207 & 0.259 & 0.288 & 0.235 & 0.247 \\
    & &Prune (S) & 0.204 & \underline{0.253} & 0.298 & 0.235 & 0.248 \\
    & &Decomp. & 0.204 & \textbf{0.251} & 0.296 & \textbf{0.228} & \underline{0.245} \\

    \toprule[1pt]
    \multirow{5}*{}
    &\multirow{5}*{PSNR $\uparrow$ }
    &MM-J & 26.18 & 21.55 & 21.65 & 24.13 & 23.38 \\
    & &+ MM & 26.35 & \textbf{22.25} & 21.42 & 24.89 & 23.73 \\
    & &Prune (H) & 26.35 & 21.73 & \underline{22.55} & 24.80 & 23.86 \\
    & &Prune (S) & \underline{26.62} & 21.44 & 22.43 & \underline{25.45} & \underline{23.99} \\
    & &Decomp. & \textbf{26.82} & \underline{22.11} & \textbf{22.57} & \textbf{25.46} & \textbf{24.24} \\

    \cmidrule{2-8}
    \multirow{5}*{T}
    &\multirow{5}*{SSIM $\uparrow$ } 
    &MM-J & 0.886 & 0.828 & 0.840 & 0.842 & 0.849 \\
    & &+ MM & 0.885 & \underline{0.847} & 0.837 & 0.855 & 0.856 \\
    & &Prune (H) & 0.891 & 0.838 & \textbf{0.862} & \underline{0.862} & 0.863 \\
    & &Prune (S) & \underline{0.892} & 0.837 & \underline{0.861} & \textbf{0.868} & \underline{0.865} \\
    & &Decomp. & \textbf{0.893} & \textbf{0.848} & 0.860 & \textbf{0.868} & \textbf{0.867} \\

    \cmidrule{2-8}
    \multirow{5}*{}
    &\multirow{5}*{LPIPS $\downarrow$} 
    &MM-J & 0.130 & 0.227 & \underline{0.274} & 0.168 & 0.200 \\
    & &+ MM & 0.149 & \underline{0.193} & 0.333 & 0.158 & 0.208 \\
    & &Prune (H) & \textbf{0.123} & 0.197 & \textbf{0.264} & 0.152 & \textbf{0.184} \\
    & &Prune (S) & \underline{0.129} & 0.200 & 0.278 & \textbf{0.145} & 0.188 \\
    & &Decomp. & 0.131 & \textbf{0.190} & 0.279 & \underline{0.147} & \underline{0.187} \\

    \toprule[1pt]
    \multirow{5}*{L}
    &\multirow{5}*{mIoU $\uparrow$ } 
    &MM-J & 56.0 & 26.9 & 44.9 & 52.4 & 45.1 \\
    & &+ MM & 55.9 & 29.0 & 44.3 & 51.9 & 45.3 \\
    & &Prune (H) & \underline{59.4} & \underline{30.3} & 44.6 & \underline{53.6} & 46.9 \\
    & &Prune (S) & \underline{59.4} & 28.7 & \textbf{46.3} & \underline{53.6} & \underline{47.0} \\
    & &Decomp. & \textbf{61.1} & \textbf{30.6} & \underline{46.1} & \textbf{54.7} & \textbf{48.1} \\
    \bottomrule[1.5pt]
  \end{tabular}
  }
  \vspace{-8pt}
\label{table:full_ablation}
\end{table}

\begin{table*}[htpb!]
    \centering
    \fontsize{4}{1}\selectfont
    \caption{Number of Gaussians ($\times 10^4$) for each scene in RGB-Thermal. ThermalGaussian is shortened as ``T-GS''. }   
    \vspace{-4pt}
    \resizebox{1.0\linewidth}{!}{
        \begin{tabular}{cccccccccccc}
        \toprule[0.45pt]
        \multicolumn{1}{c}{Method}
        &\multicolumn{1}{c}{Dim}
        &\multicolumn{1}{c}{DS}
        &\multicolumn{1}{c}{Ebk}
        &\multicolumn{1}{c}{RB}
        &\multicolumn{1}{c}{Trk}
        &\multicolumn{1}{c}{RK}
        &\multicolumn{1}{c}{Bldg}
        &\multicolumn{1}{c}{II}
        &\multicolumn{1}{c}{Pt}
        &\multicolumn{1}{c}{LS}
        &\multicolumn{1}{c}{Avg.} \\
    
        \toprule[0.25pt]
        T-GS & 32.2 & 27.0 & 19.8 & 9.3 & 27.6 & 43.3 & 66.5 & 45.9 & 20.0 & 35.7 & 32.7 \\
        \textbf{MMOne} & \textbf{7.5} & \textbf{5.5} & \textbf{10.4} & \textbf{4.4} & \textbf{9.8} & \textbf{13.6} & \textbf{23.2} & \textbf{14.4} &\textbf{ 8.4} & \textbf{12.2} & \textbf{12.2} \\
        \bottomrule[0.45pt]
        \end{tabular}
    }
    \vspace{-4pt}
\label{table:thermal_number}
\end{table*}

\begin{figure}[htpb!]
    \centering
    \includegraphics[width=1\linewidth]{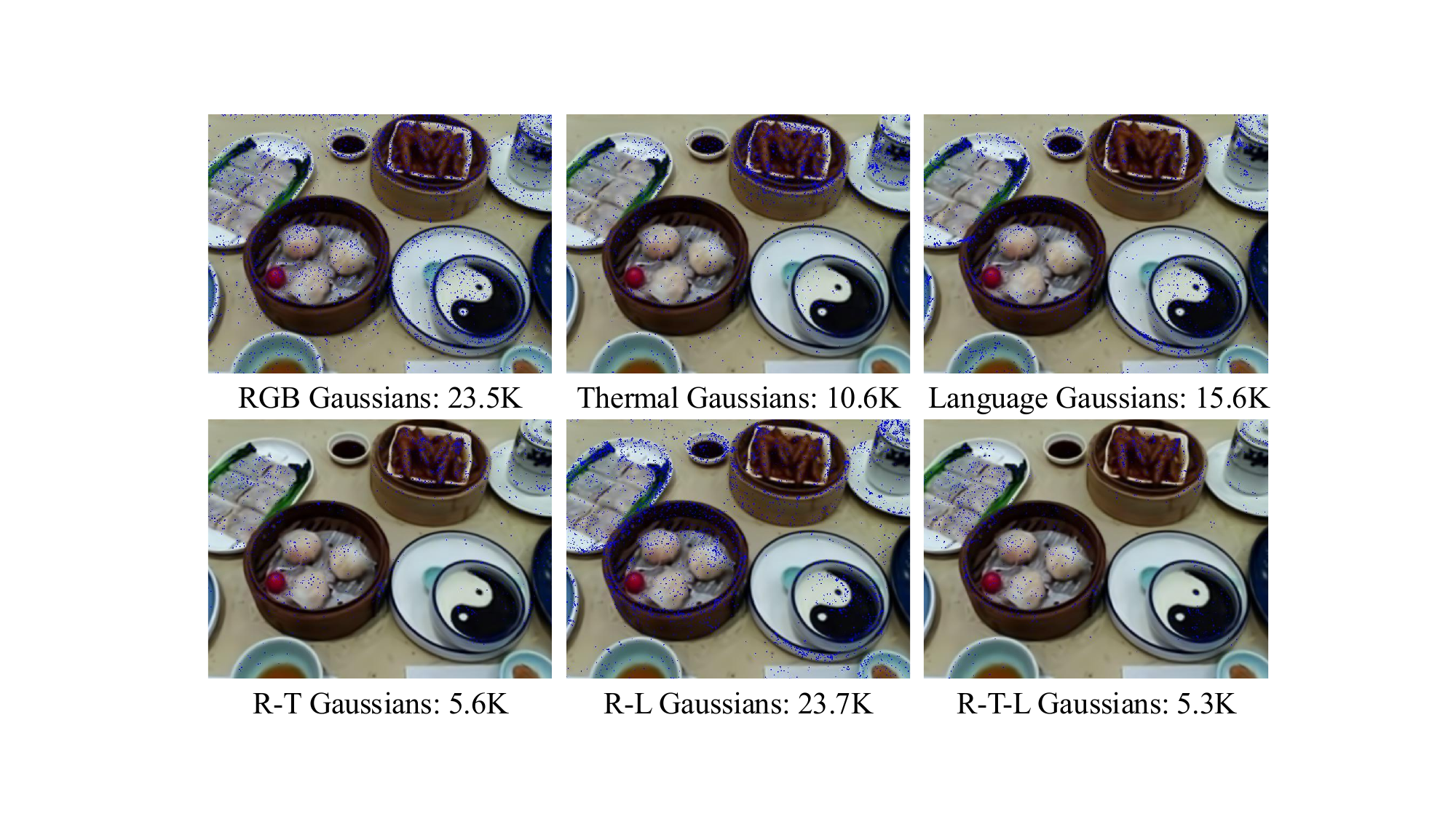}
    \caption{ \textbf{Gaussian Distributions.} The total number of Gaussians is 89.5K. ``T-L Gaussians'' are omitted for visual clarity. }  
    \label{fig:Gaussian_quantities}
    \vspace{-8pt}
\end{figure}

We also present the complete ablation results for each scene in \cref{table:full_ablation}. Our full method (Decomp.) consistently outperforms other approaches across most scenes, demonstrating the effectiveness of our multimodal decomposition mechanism. Moreover, the proposed ``Soft Prune'' method consistently surpasses ``Hard Prune'' in most scenes, highlighting its advantage in mitigating conflicts associated with pruning entire Gaussians. While our modality modeling module serves as the foundation for the multimodal decomposition mechanism, its performance remains suboptimal without the disentangling of modalities, due to the varying levels of granularity among them.

\begin{table}[htpb]
    \centering
    \caption{Number of Gaussians ($\times 10^4$) in RGB-Language. }
    \vspace{-4pt}
    \resizebox{1.0\linewidth}{!}{
    \begin{tabular}{cccccc}
    \toprule[1pt]
    \multicolumn{1}{c}{Method}
    &\multicolumn{1}{c}{Figurines}
    &\multicolumn{1}{c}{Ramen}
    &\multicolumn{1}{c}{Teatime}
    &\multicolumn{1}{c}{Kitchen}
    &\multicolumn{1}{c}{Avg.} \\

    \toprule[0.6pt]
    LS* & 92.2 & 58.6 & 182.0 & 168.0 & 125.2 \\
    LS-J & 55.1 & 31.1 & 119.0 & 105.0 & 77.6 \\
    \textbf{MMOne} & \textbf{29.5} & \textbf{16.4} & \textbf{28.9} & \textbf{42.9} & \textbf{29.4} \\
    \bottomrule[1pt]
    \end{tabular}
  }
  \vspace{-6pt}
\label{table:lerf_number}
\end{table}

\section{Additional Qualitative Results}

To analyze the distributions of multimodal and single-modal Gaussians, we use the ``Dimsum'' scene as an example. As shown in \cref{fig:Gaussian_quantities}, different modalities require varying number of Gaussians.

\begin{figure*}[htpb]
    \centering
    \includegraphics[width=1\linewidth]{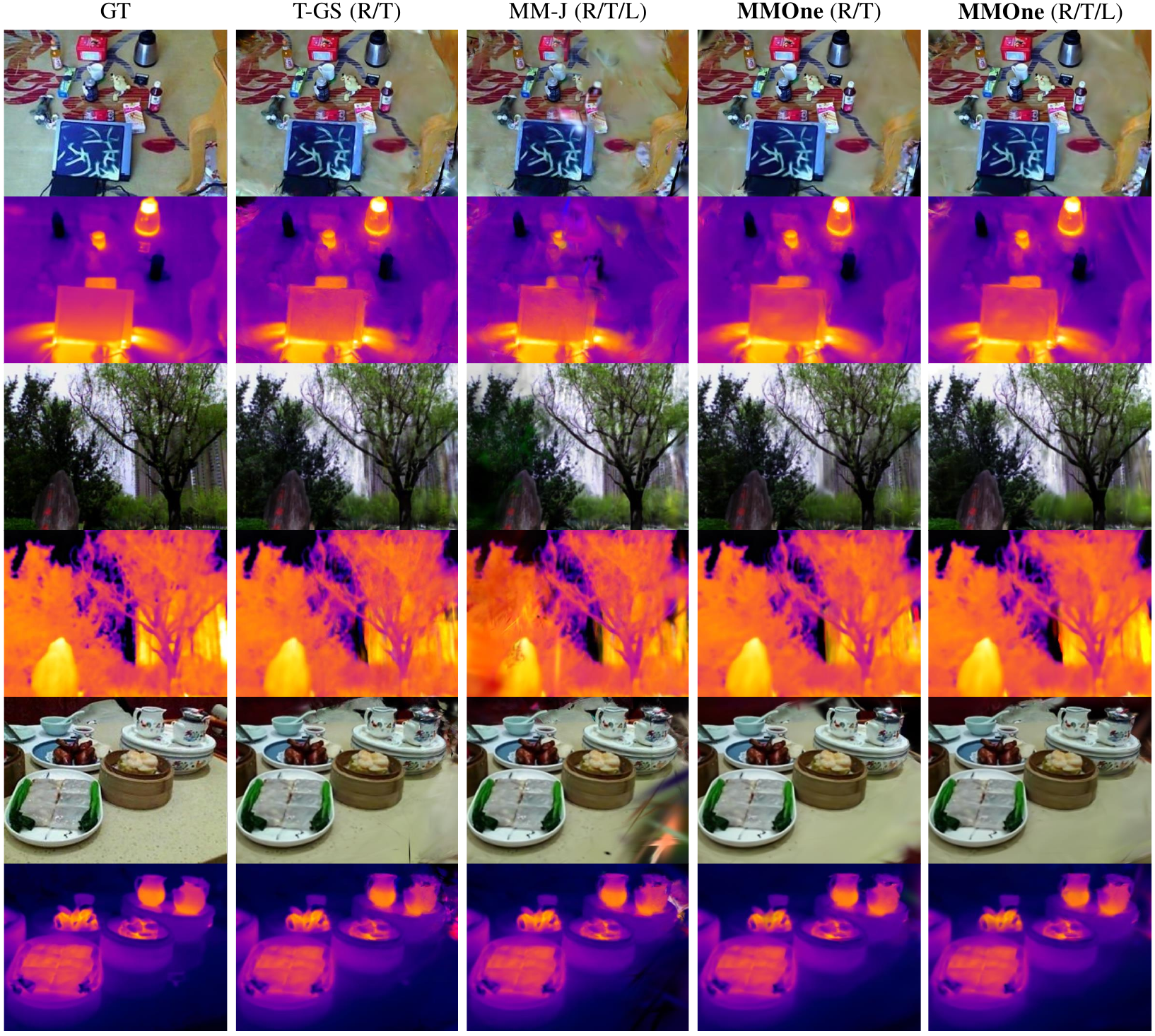}
    \caption{Qualitative results of the modality conflicts caused by the introduction of language. ``T-GS'' refers to ThermalGaussian and ``MM-J'' denotes our RGB-Thermal-Language joint training baseline. }  
    \label{fig:modality_conflicts}
\end{figure*}

We also present the qualitative results of modality conflicts in \cref{fig:modality_conflicts}. Both ``T-GS'' and ``MM-J'' refer to joint training of multiple modalities with a shared opacity. The rendering results of ``MM-J'' show significant blurring, which severely degrades the quality of both RGB and thermal renderings. This suggests that, without our proposed modality decomposition, modality conflicts become more pronounced as the number of modalities increases. This observation aligns with our intuition, as different modalities possess distinct properties. In contrast, our methods, trained on two or three modalities, consistently deliver superior results. Notably, the introduction of the language modality does not degrade the performance of RGB and thermal modalities, due to our modality modeling module and multimodal decomposition mechanism, which ensure scalability to additional modalities.

\section{Additional Quantitative Results}

We present the number of Gaussians in the RGB-Thermal and RGB-Language experiments to further highlight the effectiveness of our method in achieving a compact representation. As shown in \cref{table:thermal_number}, our method uses approximately one-third of the Gaussians utilized by ThermalGaussian. Similarly, \cref{table:lerf_number} demonstrates that our method employs only 25\% of the Gaussians used by LangSplat and 40\% of those used by the joint training baseline modified from LangSplat. These results underscore that our multimodal decomposition mechanism effectively eliminates redundant Gaussians, leading to a more compact and efficient scene representation.

For RGB-Language experiments, we additionally include Feature-3DGS as another joint training baseline with a different language rasterizer. As shown in \cref{table:feature3dgs_quantitative}, due to modality conflicts, the performance of ``F-GS'' and ``LS-J'' drops 0.7\%mIoU and 0.8dB for language and RGB, respectively. In contrast, 9.0\%mIoU and 0.3dB improvements are achieved by our MMOne.

To further demonstrate the benefits of incorporating thermal information, we conduct additional RGB-Language experiments on the ``Dimsum'' scene. As shown in \cref{table:R_L}, the inclusion of thermal data leads to improvements in both RGB rendering quality and open-vocabulary segmentation accuracy, highlighting its effectiveness in enhancing multimodal scene understanding.

\begin{table}[htpb!]
    \caption{ Additional quantitative comparisons on RGB-Language. Feature-3DGS is shortened as ``F-GS''.
    } 
    \centering
    \small
    \resizebox{1.0\linewidth}{!}{
    \begin{tabular}{lccccc}  
        \toprule[0.8pt]
        \multirow{2}{1cm}{Method}
        &\multicolumn{3}c{RGB}
        &\multicolumn{2}c{Lang} \\
  
        \cmidrule(lr){2-4} \cmidrule(lr){5-6}
        &\multicolumn{1}{c}{PSNR $\uparrow$}
        &\multicolumn{1}{c}{SSIM $\uparrow$}
        &\multicolumn{1}{c}{LPIPS $\downarrow$}
        &\multicolumn{1}{c}{mIoU $\uparrow$}
        &\multicolumn{1}{c}{acc $\uparrow$} \\

        \toprule[0.6pt]
        LS* & 24.02 & \textbf{0.854} & \textbf{0.220} & 47.6 & 72.4 \\
        
        F-GS & \underline{24.16} & \underline{0.851} & \underline{0.232} & 46.9 & 71.7 \\

        LS-J & 23.23 & 0.837 & 0.257 & \underline{55.3} & \underline{73.5} \\

        \textbf{MMOne} & \textbf{24.35} & \underline{0.851} & 0.244 & \textbf{56.6} & \textbf{76.5} \\

    \bottomrule[0.8pt]
    \end{tabular}
    }
    \vspace{-8pt}
\label{table:feature3dgs_quantitative}
\end{table}

\begin{table}[htpb!]
    \caption{ Quantitative comparisons between RGB-Language and RGB-Thermal-Language.  }
    \centering
    \small
    \resizebox{1.0\linewidth}{!}{
    \begin{tabular}{lcccc}  
        \toprule[0.8pt]
        \multirow{2}{1.2cm}{Method}
        &\multicolumn{3}c{RGB} 
        &\multicolumn{1}c{Lang} \\
  
        \cmidrule(lr){2-4} \cmidrule(lr){5-5}
        &\multicolumn{1}{c}{PSNR $\uparrow$}
        &\multicolumn{1}{c}{SSIM $\uparrow$}
        &\multicolumn{1}{c}{LPIPS $\downarrow$}
        &\multicolumn{1}{c}{mIoU $\uparrow$} \\

        \toprule[0.6pt]
        MMOne(R/L) & 24.46 & 0.861 & \textbf{0.204} & 57.1 \\
        
        MMOne(R/T/L) & \textbf{24.74} & \textbf{0.863} & \textbf{0.204} & \textbf{61.1} \\

    \bottomrule[0.8pt]
    \end{tabular}
    }
    \vspace{-12pt}
\label{table:R_L}
\end{table}

\begin{table}[htpb!]
    \caption{ Quantitative results of incorporating monocular depth.  }
    \centering
    \Huge
    \resizebox{1.0\linewidth}{!}{
    \begin{tabular}{lccccccc}  
        \toprule[2.5pt]
        \multirow{2}{1.2cm}{Method}
        &\multicolumn{3}c{RGB} 
        &\multicolumn{3}c{Thermal}
        &\multicolumn{1}c{Lang} \\
  
        \cmidrule(lr){2-4} \cmidrule(lr){5-7} \cmidrule(lr){8-8}
        &\multicolumn{1}{c}{PSNR $\uparrow$}
        &\multicolumn{1}{c}{SSIM $\uparrow$}
        &\multicolumn{1}{c}{LPIPS $\downarrow$}
        &\multicolumn{1}{c}{PSNR $\uparrow$}
        &\multicolumn{1}{c}{SSIM $\uparrow$}
        &\multicolumn{1}{c}{LPIPS $\downarrow$}
        &\multicolumn{1}{c}{mIoU $\uparrow$} \\

        \toprule[1pt]
        MMOne(R/T/L) & 24.74 & \textbf{0.863} & \textbf{0.204} & 26.82 & \textbf{0.893} & 0.131 & \textbf{61.1} \\
        
        MMOne(R/T/L/D) & \textbf{24.75} & \textbf{0.863} & 0.206 & \textbf{26.84} & \textbf{0.893} & \textbf{0.130} & \textbf{61.1} \\

    \bottomrule[2.5pt]
    \end{tabular}
    }
    \vspace{-12pt}
\label{table:depth_quantitative}
\end{table}

\begin{figure}[htpb!]
    \centering
    \includegraphics[width=1\linewidth]{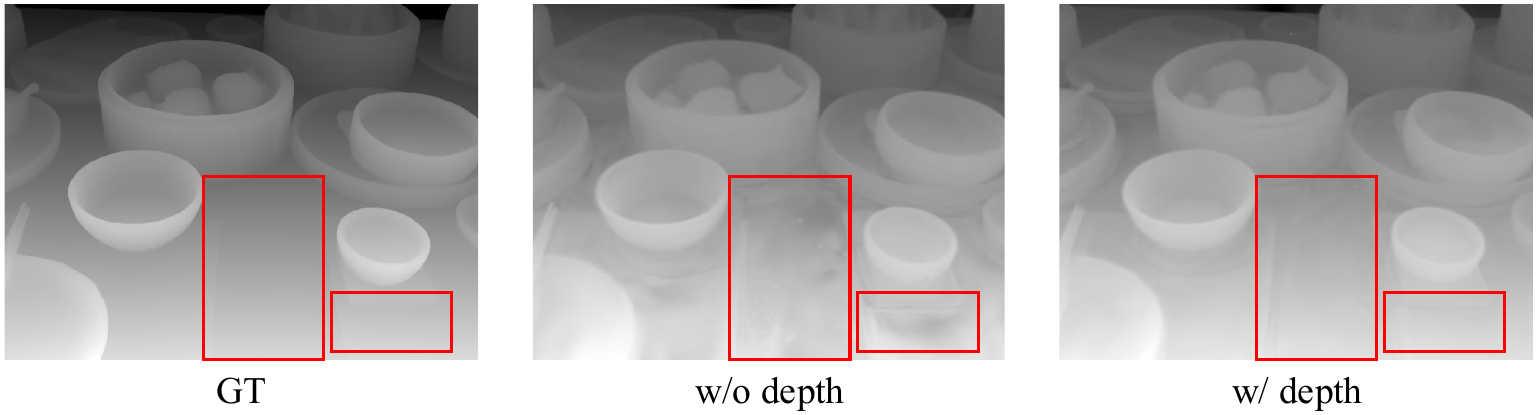}
    \vspace{-8pt}
    \caption{ Qualitative results of incorporating monocular depth.  }  
    \label{fig:depth_qualitative}
    \vspace{-8pt}
\end{figure}

\section{Additional Experiments on Scalability}

We further validate scalability by incorporating monocular depth in the ``Dimsum'' scene. The results in \cref{fig:depth_qualitative} and \cref{table:depth_quantitative} show that the rendered depth quality is enhanced, particularly on flat surfaces, without compromising the performance of RGB, thermal, and language.

\clearpage

%% file: main.bbl
\begin{thebibliography}{54}
\providecommand{\natexlab}[1]{#1}
\providecommand{\url}[1]{\texttt{#1}}
\expandafter\ifx\csname urlstyle\endcsname\relax
  \providecommand{\doi}[1]{doi: #1}\else
  \providecommand{\doi}{doi: \begingroup \urlstyle{rm}\Url}\fi

\bibitem[Barron et~al.(2021)Barron, Mildenhall, Tancik, Hedman, Martin-Brualla, and Srinivasan]{barron2021mip}
Jonathan~T Barron, Ben Mildenhall, Matthew Tancik, Peter Hedman, Ricardo Martin-Brualla, and Pratul~P Srinivasan.
\newblock {Mip-NeRF: A Multiscale Representation for Anti-Aliasing Neural Radiance Fields}.
\newblock In \emph{Proceedings of the International Conference on Computer Vision}, 2021.

\bibitem[Caron et~al.(2021)Caron, Touvron, Misra, J{\'e}gou, Mairal, Bojanowski, and Joulin]{caron2021emerging}
Mathilde Caron, Hugo Touvron, Ishan Misra, Herv{\'e} J{\'e}gou, Julien Mairal, Piotr Bojanowski, and Armand Joulin.
\newblock {Emerging Properties in Self-Supervised Vision Transformers}.
\newblock In \emph{Proceedings of the International Conference on Computer Vision}, 2021.

\bibitem[Chen et~al.(2024{\natexlab{a}})Chen, Li, Ye, Wang, Xie, Zhai, Wang, Liu, Bao, and Zhang]{chen2024pgsr}
Danpeng Chen, Hai Li, Weicai Ye, Yifan Wang, Weijian Xie, Shangjin Zhai, Nan Wang, Haomin Liu, Hujun Bao, and Guofeng Zhang.
\newblock {PGSR: Planar-based Gaussian Splatting for Efficient and High-Fidelity Surface Reconstruction}.
\newblock \emph{IEEE Transactions on Visualization and Computer Graphics}, 2024{\natexlab{a}}.

\bibitem[Chen and Wang(2024)]{chen2025survey3dgaussiansplatting}
Guikun Chen and Wenguan Wang.
\newblock {A Survey on 3D Gaussian Splatting}.
\newblock \emph{arXiv preprint arXiv:2401.03890}, 2024.

\bibitem[Chen et~al.(2024{\natexlab{b}})Chen, Shu, and Bai]{chen2024thermal3d}
Qian Chen, Shihao Shu, and Xiangzhi Bai.
\newblock {Thermal3D-GS: Physics-induced 3D Gaussians for Thermal Infrared Novel-view Synthesis}.
\newblock In \emph{Proceedings of the European Conference on Computer Vision}, 2024{\natexlab{b}}.

\bibitem[Dou et~al.(2024)Dou, Yang, Liu, Loquercio, and Owens]{dou2024tactile}
Yiming Dou, Fengyu Yang, Yi Liu, Antonio Loquercio, and Andrew Owens.
\newblock {Tactile-Augmented Radiance Fields}.
\newblock In \emph{Proceedings of the Conference on Computer Vision and Pattern Recognition}, 2024.

\bibitem[Duan et~al.(2022)Duan, Shi, and Wu]{duan2022multimodal}
Shengshun Duan, Qiongfeng Shi, and Jun Wu.
\newblock {Multimodal Sensors and ML-Based Data Fusion for Advanced Robots}.
\newblock \emph{Advanced Intelligent Systems}, 4\penalty0 (12):\penalty0 2200213, 2022.

\bibitem[Dubail et~al.(2022)Dubail, Guerrero~Pe{\~n}a, Medeiros, Aminbeidokhti, Granger, and Pedersoli]{dubail2022privacy}
Thomas Dubail, Fidel~Alejandro Guerrero~Pe{\~n}a, Heitor~Rapela Medeiros, Masih Aminbeidokhti, Eric Granger, and Marco Pedersoli.
\newblock {Privacy-Preserving Person Detection Using Low-Resolution Infrared Cameras}.
\newblock In \emph{Proceedings of the European Conference on Computer Vision}, 2022.

\bibitem[Epstein and Baker(2019)]{epstein2019scene}
Russell~A Epstein and Chris~I Baker.
\newblock {Scene Perception in the Human Brain}.
\newblock \emph{Annual Review of Vision Science}, 5\penalty0 (1):\penalty0 373--397, 2019.

\bibitem[Fang and Wang(2024{\natexlab{a}})]{fang2024mini}
Guangchi Fang and Bing Wang.
\newblock {Mini-Splatting: Representing Scenes with a Constrained Number of Gaussians}.
\newblock In \emph{Proceedings of the European Conference on Computer Vision}, 2024{\natexlab{a}}.

\bibitem[Fang and Wang(2024{\natexlab{b}})]{fang2024mini2}
Guangchi Fang and Bing Wang.
\newblock {Mini-Splatting2: Building 360 Scenes within Minutes via Aggressive Gaussian Densification}.
\newblock \emph{arXiv preprint arXiv:2411.12788}, 2024{\natexlab{b}}.

\bibitem[Gao et~al.(2022)Gao, Gao, He, Lu, Xu, and Li]{gao2022nerf}
Kyle Gao, Yina Gao, Hongjie He, Dening Lu, Linlin Xu, and Jonathan Li.
\newblock {NeRF: Neural Radiance Field in 3D Vision: A Comprehensive Review}.
\newblock \emph{arXiv preprint arXiv:2210.00379}, 2022.

\bibitem[Guo et~al.(2020)Guo, Wang, Hu, Liu, Liu, and Bennamoun]{guo2020deep}
Yulan Guo, Hanyun Wang, Qingyong Hu, Hao Liu, Li Liu, and Mohammed Bennamoun.
\newblock {Deep Learning for 3D Point Clouds: A Survey}.
\newblock \emph{IEEE Transactions on Pattern Analysis and Machine Intelligence}, 43\penalty0 (12):\penalty0 4338--4364, 2020.

\bibitem[Haque et~al.(2020)Haque, Milstein, and Fei-Fei]{haque2020illuminating}
Albert Haque, Arnold Milstein, and Li Fei-Fei.
\newblock {Illuminating the Dark Spaces of Healthcare with Ambient Intelligence}.
\newblock \emph{Nature}, 585\penalty0 (7824):\penalty0 193--202, 2020.

\bibitem[Hassan et~al.(2024)Hassan, Forest, Fink, and Mielle]{hassan2024thermonerf}
Mariam Hassan, Florent Forest, Olga Fink, and Malcolm Mielle.
\newblock {ThermoNeRF: Multimodal Neural Radiance Fields for Thermal Novel View Synthesis}.
\newblock \emph{arXiv preprint arXiv:2403.12154}, 2024.

\bibitem[Huang et~al.(2024)Huang, Yu, Chen, Geiger, and Gao]{huang20242d}
Binbin Huang, Zehao Yu, Anpei Chen, Andreas Geiger, and Shenghua Gao.
\newblock {2D Gaussian Splatting for Geometrically Accurate Radiance Fields }.
\newblock In \emph{SIGGRAPH}, 2024.

\bibitem[Intraub(2012)]{intraub2012rethinking}
Helene Intraub.
\newblock {Rethinking Visual Scene Perception}.
\newblock \emph{Wiley Interdisciplinary Reviews: Cognitive Science}, 3\penalty0 (1):\penalty0 117--127, 2012.

\bibitem[Jalil et~al.(2019)Jalil, Pascali, Leone, Martinelli, Moroni, Salvetti, and Berton]{jalil2019visible}
B Jalil, MA Pascali, GR Leone, M Martinelli, D Moroni, O Salvetti, and A Berton.
\newblock {Visible and Infrared Imaging Based Inspection of Power Installation}.
\newblock \emph{Pattern Recognition and Image Analysis}, 29\penalty0 (1):\penalty0 35--41, 2019.

\bibitem[Kerbl et~al.(2023)Kerbl, Kopanas, Leimk{\"u}hler, and Drettakis]{kerbl20233d}
Bernhard Kerbl, Georgios Kopanas, Thomas Leimk{\"u}hler, and George Drettakis.
\newblock {3D Gaussian Splatting for Real-Time Radiance Field Rendering}.
\newblock \emph{ACM Transactions on Graphics}, 42\penalty0 (4):\penalty0 139--1, 2023.

\bibitem[Kerr et~al.(2023)Kerr, Kim, Goldberg, Kanazawa, and Tancik]{kerr2023lerf}
Justin Kerr, Chung~Min Kim, Ken Goldberg, Angjoo Kanazawa, and Matthew Tancik.
\newblock {LERF: Language Embedded Radiance Fields}.
\newblock In \emph{Proceedings of the International Conference on Computer Vision}, 2023.

\bibitem[Kirillov et~al.(2023)Kirillov, Mintun, Ravi, Mao, Rolland, Gustafson, Xiao, Whitehead, Berg, Lo, et~al.]{kirillov2023segment}
Alexander Kirillov, Eric Mintun, Nikhila Ravi, Hanzi Mao, Chloe Rolland, Laura Gustafson, Tete Xiao, Spencer Whitehead, Alexander~C Berg, Wan-Yen Lo, et~al.
\newblock {Segment Anything}.
\newblock In \emph{Proceedings of the International Conference on Computer Vision}, 2023.

\bibitem[Li et~al.(2024)Li, Qin, Zou, He, Li, Dai, Zhang, and Han]{li2024langsurf}
Hao Li, Roy Qin, Zhengyu Zou, Diqi He, Bohan Li, Bingquan Dai, Dingewn Zhang, and Junwei Han.
\newblock {LangSurf: Language-Embedded Surface Gaussians for 3D Scene Understanding}.
\newblock \emph{arXiv preprint arXiv:2412.17635}, 2024.

\bibitem[Liu et~al.(2023)Liu, Zhan, Zhang, Xu, Yu, El~Saddik, Theobalt, Xing, and Lu]{liu2023weakly}
Kunhao Liu, Fangneng Zhan, Jiahui Zhang, Muyu Xu, Yingchen Yu, Abdulmotaleb El~Saddik, Christian Theobalt, Eric Xing, and Shijian Lu.
\newblock {Weakly Supervised 3D Open-vocabulary Segmentation}.
\newblock \emph{Advances in Neural Information Processing Systems}, 36:\penalty0 53433--53456, 2023.

\bibitem[Liu et~al.(2025)Liu, Guan, Zhu, Xu, Song, Li, Wang, and Yang]{liu2025efficientgs}
Wenkai Liu, Tao Guan, Bin Zhu, Luoyuan Xu, Zikai Song, Dan Li, Yuesong Wang, and Wei Yang.
\newblock {EfficientGS: Streamlining Gaussian Splatting for Large-Scale High-Resolution Scene Representation}.
\newblock \emph{IEEE MultiMedia}, 2025.

\bibitem[Lu et~al.(2025)Lu, Chen, Zhu, Qin, Lu, Zhang, Yan, and Xue]{lu2025thermalgaussian}
Rongfeng Lu, Hangyu Chen, Zunjie Zhu, Yuhang Qin, Ming Lu, Le Zhang, Chenggang Yan, and Anke Xue.
\newblock {ThermalGaussian: Thermal 3D Gaussian Splatting}.
\newblock In \emph{Proceedings of the International Conference on Learning Representations}, 2025.

\bibitem[Mallick et~al.(2024)Mallick, Goel, Kerbl, Steinberger, Carrasco, and De~La~Torre]{mallick2024taming}
Saswat~Subhajyoti Mallick, Rahul Goel, Bernhard Kerbl, Markus Steinberger, Francisco~Vicente Carrasco, and Fernando De~La~Torre.
\newblock {Taming 3DGS: High-Quality Radiance Fields with Limited Resources}.
\newblock In \emph{SIGGRAPH Asia}, 2024.

\bibitem[Mildenhall et~al.(2021)Mildenhall, Srinivasan, Tancik, Barron, Ramamoorthi, and Ng]{mildenhall2021nerf}
Ben Mildenhall, Pratul~P Srinivasan, Matthew Tancik, Jonathan~T Barron, Ravi Ramamoorthi, and Ren Ng.
\newblock {NeRF: Representing Scenes as Neural Radiance Fields for View Synthesis}.
\newblock \emph{Communications of the ACM}, 65\penalty0 (1):\penalty0 99--106, 2021.

\bibitem[Osornio-Rios et~al.(2018)Osornio-Rios, Antonino-Daviu, and de~Jesus Romero-Troncoso]{osornio2018recent}
Roque~Alfredo Osornio-Rios, Jose~Alfonso Antonino-Daviu, and Rene de Jesus Romero-Troncoso.
\newblock {Recent Industrial Applications of Infrared Thermography: A Review}.
\newblock \emph{IEEE Transactions on Industrial Informatics}, 15\penalty0 (2):\penalty0 615--625, 2018.

\bibitem[{\"O}zer et~al.(2024){\"O}zer, Weiherer, Hundhausen, and Egger]{ozer2024exploring}
Mert {\"O}zer, Maximilian Weiherer, Martin Hundhausen, and Bernhard Egger.
\newblock {Exploring Multi-modal Neural Scene Representations With Applications on Thermal Imaging}.
\newblock In \emph{Proceedings of the European Conference on Computer Vision}, 2024.

\bibitem[Peng et~al.(2025)Peng, Planche, Gao, Zheng, Choudhuri, Chen, Chen, and Wu]{peng2025d}
Qucheng Peng, Benjamin Planche, Zhongpai Gao, Meng Zheng, Anwesa Choudhuri, Terrence Chen, Chen Chen, and Ziyan Wu.
\newblock {3D Vision-Language Gaussian Splatting}.
\newblock In \emph{Proceedings of the International Conference on Learning Representations}, 2025.

\bibitem[Poggi et~al.(2022)Poggi, Ramirez, Tosi, Salti, Mattoccia, and Di~Stefano]{poggi2022cross}
Matteo Poggi, Pierluigi~Zama Ramirez, Fabio Tosi, Samuele Salti, Stefano Mattoccia, and Luigi Di~Stefano.
\newblock {Cross-Spectral Neural Radiance Fields}.
\newblock In \emph{Proceedings of the International Conference on 3D Vision}, 2022.

\bibitem[Qian et~al.(2024{\natexlab{a}})Qian, Kirschstein, Schoneveld, Davoli, Giebenhain, and Nie{\ss}ner]{qian2024gaussianavatars}
Shenhan Qian, Tobias Kirschstein, Liam Schoneveld, Davide Davoli, Simon Giebenhain, and Matthias Nie{\ss}ner.
\newblock {GaussianAvatars: Photorealistic Head Avatars with Rigged 3D Gaussians}.
\newblock In \emph{Proceedings of the Conference on Computer Vision and Pattern Recognition}, 2024{\natexlab{a}}.

\bibitem[Qian et~al.(2024{\natexlab{b}})Qian, Wang, Mihajlovic, Geiger, and Tang]{qian20243dgs}
Zhiyin Qian, Shaofei Wang, Marko Mihajlovic, Andreas Geiger, and Siyu Tang.
\newblock {3DGS-Avatar: Animatable Avatars via Deformable 3D Gaussian Splatting}.
\newblock In \emph{Proceedings of the Conference on Computer Vision and Pattern Recognition}, 2024{\natexlab{b}}.

\bibitem[Qin et~al.(2024)Qin, Li, Zhou, Wang, and Pfister]{qin2024langsplat}
Minghan Qin, Wanhua Li, Jiawei Zhou, Haoqian Wang, and Hanspeter Pfister.
\newblock {LangSplat: 3D Language Gaussian Splatting}.
\newblock In \emph{Proceedings of the Conference on Computer Vision and Pattern Recognition}, 2024.

\bibitem[Qiu et~al.(2024)Qiu, Liu, Su, and Lin]{qiu2024gls}
Jiaxiong Qiu, Liu Liu, Zhizhong Su, and Tianwei Lin.
\newblock {GLS: Geometry-aware 3D Language Gaussian Splatting}.
\newblock \emph{arXiv preprint arXiv:2411.18066}, 2024.

\bibitem[Qu et~al.(2024)Qu, Dai, Li, Lin, Cao, Zhang, and Ji]{qu2024goi}
Yansong Qu, Shaohui Dai, Xinyang Li, Jianghang Lin, Liujuan Cao, Shengchuan Zhang, and Rongrong Ji.
\newblock {GOI: Find 3D Gaussians of Interest with an Optimizable Open-vocabulary Semantic-space Hyperplane}.
\newblock In \emph{Proceedings of the International Conference on Multimedia}, 2024.

\bibitem[Radford et~al.(2021)Radford, Kim, Hallacy, Ramesh, Goh, Agarwal, Sastry, Askell, Mishkin, Clark, et~al.]{radford2021learning}
Alec Radford, Jong~Wook Kim, Chris Hallacy, Aditya Ramesh, Gabriel Goh, Sandhini Agarwal, Girish Sastry, Amanda Askell, Pamela Mishkin, Jack Clark, et~al.
\newblock {Learning Transferable Visual Models From Natural Language Supervision }.
\newblock In \emph{Proceedings of the International Conference on Machine Learning}, 2021.

\bibitem[Saputra et~al.(2021)Saputra, Lu, de~Gusmao, Wang, Markham, and Trigoni]{saputra2021graph}
Muhamad Risqi~U Saputra, Chris~Xiaoxuan Lu, Pedro Porto~B de Gusmao, Bing Wang, Andrew Markham, and Niki Trigoni.
\newblock {Graph-based Thermal-Inertial SLAM with Probabilistic Neural Networks}.
\newblock \emph{IEEE Transactions on Robotics}, 38\penalty0 (3):\penalty0 1875--1893, 2021.

\bibitem[Schonberger and Frahm(2016)]{schonberger2016structure}
Johannes~L Schonberger and Jan-Michael Frahm.
\newblock {Structure-From-Motion Revisited}.
\newblock In \emph{Proceedings of the Conference on Computer Vision and Pattern Recognition}, 2016.

\bibitem[Shi et~al.(2024)Shi, Wang, Duan, and Guan]{shi2024language}
Jin-Chuan Shi, Miao Wang, Hao-Bin Duan, and Shao-Hua Guan.
\newblock {Language Embedded 3D Gaussians for Open-Vocabulary Scene Understanding}.
\newblock In \emph{Proceedings of the Conference on Computer Vision and Pattern Recognition}, 2024.

\bibitem[Shorinwa et~al.(2024)Shorinwa, Tucker, Smith, Swann, Chen, Firoozi, Kennedy, and Schwager]{shorinwa2024splatmover}
Olaolu Shorinwa, Johnathan Tucker, Aliyah Smith, Aiden Swann, Timothy Chen, Roya Firoozi, Monroe~David Kennedy, and Mac Schwager.
\newblock {Splat-MOVER: Multi-Stage, Open-Vocabulary Robotic Manipulation via Editable Gaussian Splatting}.
\newblock In \emph{Proceedings of the Conference on Robot Learning}, 2024.

\bibitem[Sitzmann et~al.(2020)Sitzmann, Martel, Bergman, Lindell, and Wetzstein]{sitzmann2020implicit}
Vincent Sitzmann, Julien Martel, Alexander Bergman, David Lindell, and Gordon Wetzstein.
\newblock {Implicit Neural Representations with Periodic Activation Functions}.
\newblock \emph{Advances in Neural Information Processing Systems}, 33:\penalty0 7462--7473, 2020.

\bibitem[Swann et~al.(2024)Swann, Strong, Do, Camps, Schwager, and Kennedy]{swann2024touch}
Aiden Swann, Matthew Strong, Won~Kyung Do, Gadiel~Sznaier Camps, Mac Schwager, and Monroe Kennedy.
\newblock {Touch-GS: Visual-Tactile Supervised 3D Gaussian Splatting}.
\newblock In \emph{Proceedings of the International Conference on Intelligent Robots and Systems}, 2024.

\bibitem[Tancik et~al.(2023)Tancik, Weber, Ng, Li, Yi, Wang, Kristoffersen, Austin, Salahi, Ahuja, et~al.]{tancik2023nerfstudio}
Matthew Tancik, Ethan Weber, Evonne Ng, Ruilong Li, Brent Yi, Terrance Wang, Alexander Kristoffersen, Jake Austin, Kamyar Salahi, Abhik Ahuja, et~al.
\newblock {Nerfstudio: A Modular Framework for Neural Radiance Field Development}.
\newblock In \emph{SIGGRAPH}, 2023.

\bibitem[Turkulainen et~al.(2025)Turkulainen, Ren, Melekhov, Seiskari, Rahtu, and Kannala]{turkulainen2024dnsplatter}
Matias Turkulainen, Xuqian Ren, Iaroslav Melekhov, Otto Seiskari, Esa Rahtu, and Juho Kannala.
\newblock {DN-Splatter: Depth and Normal Priors for Gaussian Splatting and Meshing}.
\newblock In \emph{Proceedings of the Winter Conference on Applications of Computer Vision}, 2025.

\bibitem[Wallace(2004)]{wallace2004development}
Mark~T Wallace.
\newblock {The Development of Multisensory Processes}.
\newblock \emph{Cognitive Processing}, 5:\penalty0 69--83, 2004.

\bibitem[Wu et~al.(2024{\natexlab{a}})Wu, Yi, Fang, Xie, Zhang, Wei, Liu, Tian, and Wang]{wu20244d}
Guanjun Wu, Taoran Yi, Jiemin Fang, Lingxi Xie, Xiaopeng Zhang, Wei Wei, Wenyu Liu, Qi Tian, and Xinggang Wang.
\newblock {4D Gaussian Splatting for Real-Time Dynamic Scene Rendering}.
\newblock In \emph{Proceedings of the Conference on Computer Vision and Pattern Recognition}, 2024{\natexlab{a}}.

\bibitem[Wu et~al.(2024{\natexlab{b}})Wu, Meng, Li, Wu, Shi, Cheng, Zhao, Feng, Ding, Wang, et~al.]{wu2024opengaussian}
Yanmin Wu, Jiarui Meng, Haijie Li, Chenming Wu, Yahao Shi, Xinhua Cheng, Chen Zhao, Haocheng Feng, Errui Ding, Jingdong Wang, et~al.
\newblock {OpenGaussian: Towards Point-Level 3D Gaussian-based Open Vocabulary Understanding}.
\newblock \emph{Advances in Neural Information Processing Systems}, 37:\penalty0 19114--19138, 2024{\natexlab{b}}.

\bibitem[Xie et~al.(2024)Xie, Zong, Qiu, Li, Feng, Yang, and Jiang]{xie2024physgaussian}
Tianyi Xie, Zeshun Zong, Yuxing Qiu, Xuan Li, Yutao Feng, Yin Yang, and Chenfanfu Jiang.
\newblock {PhysGaussian: Physics-Integrated 3D Gaussians for Generative Dynamics}.
\newblock In \emph{Proceedings of the Conference on Computer Vision and Pattern Recognition}, 2024.

\bibitem[Yan et~al.(2024)Yan, Lin, Zhou, Wang, Sun, Zhan, Lang, Zhou, and Peng]{yan2024street}
Yunzhi Yan, Haotong Lin, Chenxu Zhou, Weijie Wang, Haiyang Sun, Kun Zhan, Xianpeng Lang, Xiaowei Zhou, and Sida Peng.
\newblock {Street Gaussians: Modeling Dynamic Urban Scenes with Gaussian Splatting}.
\newblock In \emph{Proceedings of the European Conference on Computer Vision}, 2024.

\bibitem[Yu et~al.(2024)Yu, Sattler, and Geiger]{yu2024gaussian}
Zehao Yu, Torsten Sattler, and Andreas Geiger.
\newblock {Gaussian Opacity Fields: Efficient Adaptive Surface Reconstruction in Unbounded Scenes}.
\newblock \emph{ACM Transactions on Graphics}, 43\penalty0 (6):\penalty0 1--13, 2024.

\bibitem[Zhang et~al.(2020)Zhang, Yang, He, and Deng]{zhang2020multimodal}
Chao Zhang, Zichao Yang, Xiaodong He, and Li Deng.
\newblock {Multimodal Intelligence: Representation Learning, Information Fusion, and Applications}.
\newblock \emph{IEEE Journal of Selected Topics in Signal Processing}, 14\penalty0 (3):\penalty0 478--493, 2020.

\bibitem[Zhou et~al.(2024{\natexlab{a}})Zhou, Chang, Jiang, Fan, Zhu, Xu, Chari, You, Wang, and Kadambi]{zhou2024feature}
Shijie Zhou, Haoran Chang, Sicheng Jiang, Zhiwen Fan, Zehao Zhu, Dejia Xu, Pradyumna Chari, Suya You, Zhangyang Wang, and Achuta Kadambi.
\newblock {Feature 3DGS: Supercharging 3D Gaussian Splatting to Enable Distilled Feature Fields}.
\newblock In \emph{Proceedings of the Conference on Computer Vision and Pattern Recognition}, 2024{\natexlab{a}}.

\bibitem[Zhou et~al.(2024{\natexlab{b}})Zhou, Lin, Shan, Wang, Sun, and Yang]{zhou2024drivinggaussian}
Xiaoyu Zhou, Zhiwei Lin, Xiaojun Shan, Yongtao Wang, Deqing Sun, and Ming-Hsuan Yang.
\newblock {DrivingGaussian: Composite Gaussian Splatting for Surrounding Dynamic Autonomous Driving Scenes}.
\newblock In \emph{Proceedings of the Conference on Computer Vision and Pattern Recognition}, 2024{\natexlab{b}}.

\end{thebibliography}
